\documentclass{article}

\usepackage{microtype}
\usepackage{graphicx}
\usepackage{subfigure}
\usepackage{booktabs} 

\usepackage{hyperref}

\usepackage[accepted]{icml2024}

\usepackage{amsmath}
\usepackage{amssymb}
\usepackage{mathtools}
\usepackage{amsthm}

\usepackage{bbm}
\usepackage{wrapfig}
\usepackage{graphicx}
\usepackage{multirow}

\usepackage[capitalize,noabbrev]{cleveref}

\theoremstyle{plain}

\theoremstyle{definition}

\theoremstyle{remark}

\usepackage[textsize=tiny]{todonotes}

\icmltitlerunning{Retrieval-Augmented Score Distillation for Text-to-3D Generation}

\begin{document}

\twocolumn[
\icmltitle{Retrieval-Augmented Score Distillation for Text-to-3D Generation}



\icmlsetsymbol{equal}{*}

\begin{icmlauthorlist}
\icmlauthor{Junyoung Seo}{equal,univ}
\icmlauthor{Susung Hong}{equal,univ}
\icmlauthor{Wooseok Jang}{equal,univ}
\icmlauthor{Inès Hyeonsu Kim}{univ} \\
\icmlauthor{Minseop Kwak}{univ}
\icmlauthor{Doyup Lee}{comp}
\icmlauthor{Seungryong Kim}{univ} \\
\end{icmlauthorlist}

\icmlaffiliation{univ}{Korea Univeristy, Seoul, Korea}
\icmlaffiliation{comp}{Runway, New York, USA}

\icmlcorrespondingauthor{Seungryong Kim}{seungryong\_kim@korea.ac.kr}
\icmlcorrespondingauthor{Doyup Lee}{doyup@runwayml.com}

\icmlkeywords{Machine Learning}

\vskip 0.3in
]



\printAffiliationsAndNotice{\icmlEqualContribution} 

\setlength\intextsep{0pt}
\begin{figure*}[t]
  \centering
\includegraphics[width=1\textwidth]{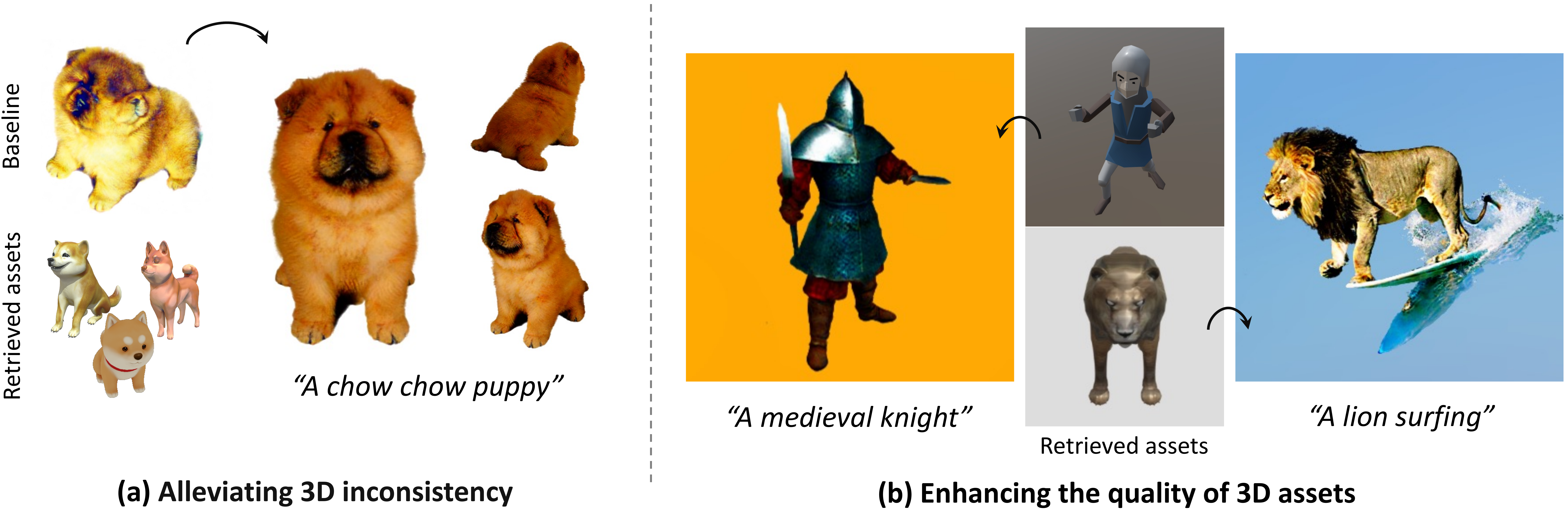}
\vspace{-20pt}
\caption{\textbf{Our framework enables to high-quality generation of 3D contentsby leveraging retrieved assets from external databases}, achieving significant enhancement of robust geometric consistency, as demonstrated in (a), and also enhancement of detail and fidelity, as shown in (b), without being bounded by the textural quality of the 3D assets.}
  \vspace{-5pt}
  \label{fig:teaser}
\end{figure*}

\begin{abstract}                       
Text-to-3D generation has achieved significant success by incorporating powerful 2D diffusion models, but insufficient 3D prior knowledge also leads to the inconsistency of 3D geometry. Recently, since large-scale multi-view datasets have been released, fine-tuning the diffusion model on the multi-view datasets becomes a mainstream to solve the 3D inconsistency problem. However, it has confronted with fundamental difficulties regarding the limited quality and diversity of 3D data, compared with 2D data. To sidestep these trade-offs, we explore a retrieval-augmented approach tailored for score distillation, dubbed ReDream. We postulate that both expressiveness of 2D diffusion models and geometric consistency of 3D assets can be fully leveraged by employing the semantically relevant assets directly within the optimization process. To this end, we introduce novel framework for retrieval-based quality enhancement in text-to-3D generation. We leverage the retrieved asset to incorporate its geometric prior in the variational objective and adapt the diffusion model's 2D prior toward view consistency, achieving drastic improvements in both geometry and fidelity of generated scenes. We conduct extensive experiments to demonstrate that ReDream exhibits superior quality with increased geometric consistency. Project page is available at \textbf{\textnormal{\url{https://ku-cvlab.github.io/ReDream/}}}.

\end{abstract}
\section{Introduction}


Text-to-3D generation has emerged as an important application that enables non-experts to easily create 3D contents. The conventional approaches for text-to-3D train a generative model directly on 3D data from scratch~\cite{wu2016learning, chen2019text2shape, zhou20213d}. However, their performance is limited due to the insufficient quality and diversity of 3D datasets compared with 2D datasets. 


The seminal works for text-to-3D~\cite{poole2022dreamfusion,wang2023score} have introduced Score Distillation Sampling (SDS) to leverage the 2D diffusion models trained on large-scale images~\cite{laion5b}. 
Given a text prompt, SDS-based frameworks~\cite{chen2023fantasia3d, seo2023let, lin2023magic3d,wang2023prolificdreamer} directly optimize a Neural Radiance Field (NeRF)~\cite{mildenhall2021nerf} by distilling the scores of text-to-image (T2I) diffusion models through the rendered views of the optimizing NeRF.
Exploiting the capability of T2I models to synthesize high-quality images~\cite{rombach2022high,saharia2022photorealistic}, SDS-based frameworks have generated high-fidelity 3D models even without 3D datasets.  
However, the generated scenes often suffer from artifacts and geometric inconsistencies due to the lack of knowledge on 3D geometry~\cite{armandpour2023re, hong2023debiasing}.

Recent approaches~\cite{zero123,mvdream} focus on fine-tuning 2D diffusion models on a large 3D dataset for novel view synthesis. 
Existing approaches~\cite{zero123,mvdream} modify and fine-tune a T2I model on Objaverse~\cite{deitke2023objaverse,deitke2023objaverse_xl} to incorporate 3D awareness into its parameters for synthesizing novel multi-views. However, compared with 2D images, the insufficiency of high-quality 3D data has consequence of severely limiting and confining the style and fidelity of the generated novel views. For example, MVDream, trained on Objaverse, undergoes a cartoonish style shift~\cite{shi2023zero123++} , hindering the model from generating photorealistc 3D textures, and Zero123 shows drastically weakened performance when photorealistic images are given as input. 



To address these issues, we propose a novel \textit{retrieval-augmented} framework, ReDream, for text-to-3D generation to leverage 3D data information without full fine-tuning of 2D diffusion models. Our key motivation is that 3D assets, which are semantically aligned with a given text, become a minimal yet effective guidance of 3D geometries for SDS-based approaches. 
Then, ReDream can largely maintain the quality of the pre-trained 2D diffusion model, but also provide an effective geometric prior.


Specifically, by interpreting each 3D scene represented by NeRF as sampled particles from a variational distribution, we show that retrieved assets can form a powerful initial variational distribution that incorporates geometric robustness and semantic relevance, grounding the generation process in these desirable qualities that text-to-3D generated scenes oftentimes lack. We also demonstrate that the retrieved assets can be leveraged for lightweight adaptation of 2D prior models, gearing the model towards more view-consistent 3D generation. These elegant and simple approaches effectively facilitates generation of high-quality 3D assets with added controllability and negligible training cost.


%

Our main contributions are summarized as follows:
\begin{itemize}
\vspace{-8pt}
\item We present an intuitive yet feasible framework, \textbf{ReDream}, that effectively integrates the retrieval module with SDS-based frameworks for text-to-3D generation.
\item Our framework can exploit both the geometric information of 3D assets and the capability of T2I models to synthesize high-fidelity images without the need of full training of the model parameters.
\item We introduce a lightweight approach that significantly reduces viewpoint bias in 2D prior models, which has been plaguing text-to-3D generation.
\item We conduct extensive experiments to demonstrate that our proposed methods consistently improve the generation quality and analyze how the retrieval-augmentation affects the 3D generation process.
\end{itemize}

\section{Related work}
\paragraph{Generative novel view synthesis.}

Generative models have been employed to learn a multi-view geometry to synthesize novel views of a 3D scene~\cite{wiles2020synsin, rombach2021geometry}. When given a single reference view, \cite{chan2023generative} estimate its 3D volume to condition a model for generating novel views. This process involves incorporating a cross-view attention in a diffusion model to align the correspondences between novel and reference views~\cite{sparsefusion,3dim}. Zero123~\cite{zero123} adapts the Stable Diffusion model~\cite{ldm} to fine-tune its entire parameters on Objaverse datasets~\cite{deitke2023objaverse_xl,deitke2023objaverse} for generating novel views of 3D objects in the open domain. However, these previous approaches face limitations in fidelity due to the scarcity of high-quality 3D data, which often requires the laborious and specialized work of experts. Additionally, MVDream~\cite{mvdream} concurrently proposes a multi-view diffusion model by fine-tuning the Stable Diffusion model.

\vspace{-5px}
\paragraph{Text-to-3D generation with score distillation.}
DreamFusion~\cite{poole2022dreamfusion} introduced a novel method known as Score Distillation Sampling (SDS) for generating 3D content without relying on 3D data. This method involves optimizing a 3D representation, such as Neural Radiance Fields (NeRF)~\cite{mildenhall2021nerf}, by distilling the prior knowledge of diffusion models to synthesize high-fidelity images. Concurrently, related studies~\cite{wang2023score} have derived similar loss functions using SDS. Following this, subsequent research~\cite{metzer2023latent, tsalicoglou2023textmesh} has consistently improved text-to-3D generation based on the SDS framework.
Other developments in this area include Magic3D~\cite{lin2023magic3d}, which utilizes DMTet~\cite{dmtet} within a coarse-to-fine pipeline to enhance the quality of 3D representation. Fantasia3D~\cite{chen2023fantasia3d} introduces a two-stage framework to separate geometry and texture in 3D content creation. ProlificDreamer~\cite{wang2023prolificdreamer} employs a particle-optimization framework for Variational Score Distillation (VSD), significantly improving the fidelity of generated textures.
However, a common challenge faced by these methods, which do not use 3D training data, is the issue of 3D inconsistency. This often results in the unrealistic geometry of the generated contents, highlighting a key area for further improvement in the field of 3D content generation.

\paragraph{Retrieval-augmented generative models.}
Retrieval-augmented approaches utilize an external database to adapt a generative model for diverse tasks without fine-tuning whole parameters on large-scale data.
For example, RETRO~\cite{retro} adapts a large language model for exploiting the external databases and achieves high performances without increasing its parameters.
For the task of image synthesis, retrieval-augmented methods have been applied to GANs~\cite{retrievegan,icgan} and diffusion models~\cite{blattmann2022retrieval,sheynin2022knn,chen2022re}, while adapting the models for synthesizing unseen styles such as artistic images~\cite{artistic_rdm}.
Since retrieval-augmentation is effective when the data scale is insufficient to train the model parameters, ~\cite{zhang2023remodiffuse} and ~\cite{he2023animate} integrate a motion-retrieval module with diffusion models to synthesize motion sequences and videos, respectively.

\section{Background: Score distillation sampling} 
Score distillation sampling (SDS)~\cite{poole2022dreamfusion} has been proposed as a method to leverage text-to-image diffusion models~\cite{saharia2022photorealistic, rombach2022high} originally trained on text-paired image datasets for generation of 3D objects. Specifically, 3D scene $\theta$, a differentiable representation such as NeRF~\cite{mildenhall2021nerf}, is optimized so that its renderings at various camera poses follow probability density $p_\phi(x|c)$ which is the 2D distribution conditioned on input text tokens $c$. The score of this distribution is approximated by the diffusion model $\epsilon_\phi$, and the practical update rule is derived as follows:
\begin{equation}
        \nabla_\theta \mathcal{L}_{\textrm{SDS}}=-\mathbb{E}_{t,\epsilon,\psi}\Big[w(t)\big(\epsilon_{\phi}(x_t|c,t) - \epsilon\big)\frac{\partial g(\theta,\psi)}{\partial \theta}\Big],
\label{eq:sds_update_rule}
\end{equation}
where $w(t)$ and $x_t$ are a weighting function and a perturbed image of $x$ with a noise level $t$, and $\epsilon$ is a corresponding Gaussian noise. $g(\cdot)$ and $\psi$ are the differentiable renderer and the camera pose, respectively. 

Variational score distillation (VSD)~\cite{wang2023prolificdreamer} further generalizes this sampling technique by interpreting it as the variational problem of fiding the distribution $\gamma$ which is represented by the particles $\theta$. Specifically, the variational distribution $ q^\gamma\big(x_t|c,x=g(\theta,\psi)\big)$ represents an implicit distribution of rendered images. The VSD framework establishes this implicit relationship by the denoising score matching process leveraging low-rank adaptation (LoRA)~\cite{ryu2023low}, resulting in following approximation: $\nabla_{x_t}q^\gamma\big(x_t|c,x=g(\theta,\psi)\big) \approx -\epsilon_{\phi,\zeta}(x_t|c,t,\psi)/{\sigma_t}$, where $\zeta$ represents a set of parameters for LoRA of the diffusion model. As a consequence, the resulting updating direction corresponds to:
\begin{equation}
\begin{aligned}
        \nabla_\theta \mathcal{L}_\mathrm{VSD}=&-\mathbb{E}_{t,\epsilon,\psi}\Big[w(t)\big(\epsilon_{\phi}(x_t|c,t) \\&- \epsilon_{\phi,\zeta}(x_t|c,t,\psi)\big)\frac{\partial g(\theta,\psi)}{\partial \theta}\Big].
\end{aligned}
\label{eq:sds_ode}
\end{equation}
For the detailed explanation on the background, please refer to Appendix~\ref{sec:theory}.

\begin{figure*}[t]
\center
\includegraphics[width=1.0\textwidth]{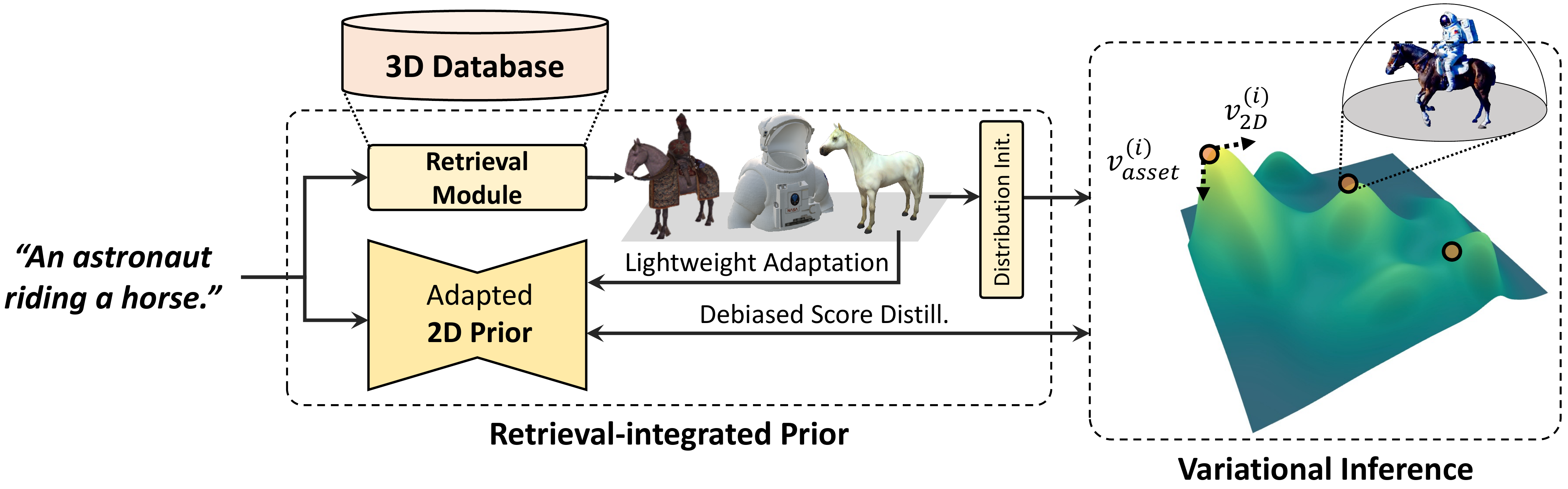}
\vspace{-10px}
\caption{\textbf{Overview.} Given a prompt $c$, we retrieve the nearest neighboring assets from the 3D database. With these assets, 
we perform initialization of an variational distribution for incorporation of robust 3D geometric prior, as well as conducting lightweight adaptation of 2D prior model for equalize probability density across viewpoints.}
\vspace{-5px}
\label{fig:overview}
\end{figure*}

\begin{figure}[t]
  \centering
\includegraphics[width=1.0\linewidth]{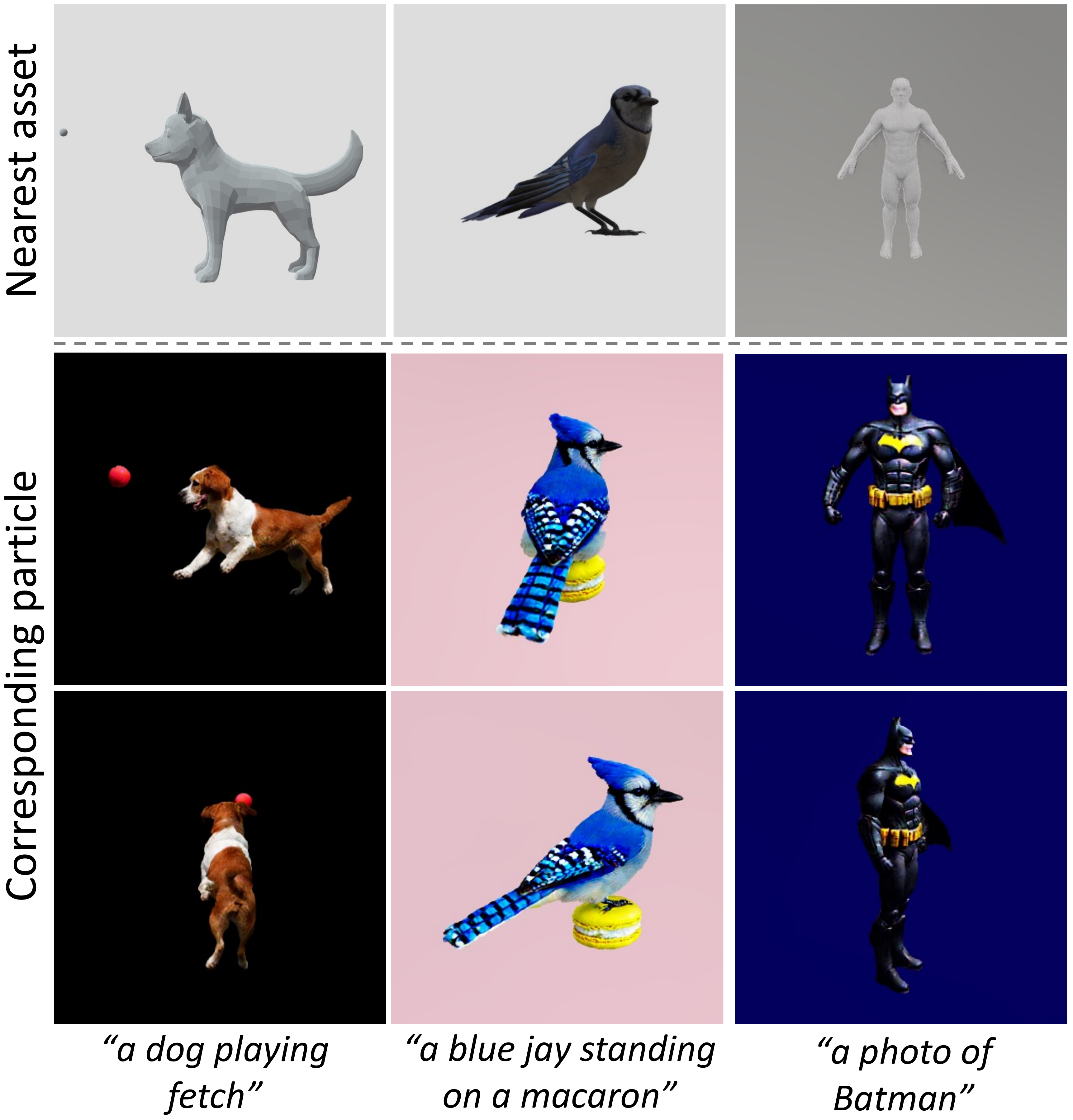}
  \vspace{-10pt}
\caption{\textbf{Generated results and corresponding nearest asset}. The first row shows the first nearest neighbor from the retrieved assets, with the renderings of corresponding particles from the given texts displayed below.}
  \vspace{-20pt}
  \label{fig:nearest}
\end{figure}

\begin{figure*}[t]
\center
\vspace{-5px}
\includegraphics[width=1.0\textwidth]{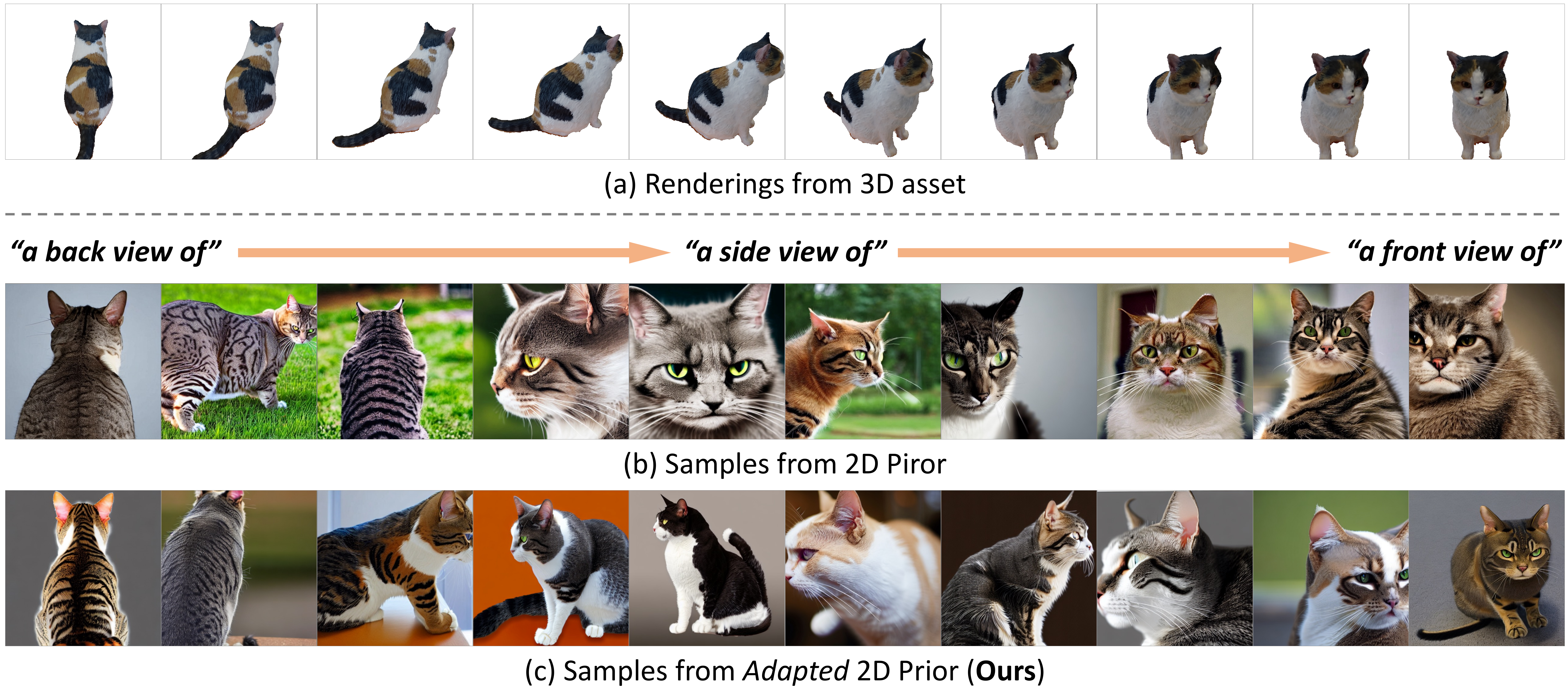}
\vspace{-10px}
\caption{\textbf{Lightweight adaptation of 2D diffusion models.} We compare the effectiveness of the adaptation with given rendering from a 3D asset in (a). We linearly interpolate a text embedding from ``\textit{a back view of an angry cat}'' to ``\textit{a front view of an angry cat}'' through ``\textit{side view}''. (b) 2D samples from the prior model. (c) 2D samples from the adapted prior model with learned view prefixes. Compared with (b). The samples from adapted 2D prior in (c) reflect a variety of viewpoints, \textbf{not biased towards a single viewpoint}.}
\vspace{-10px}
\label{fig:adaptation}
\end{figure*}

\vspace{-10px}
\section{Retrieval-augmented score distillation}
\label{sec:method}
\subsection{Motivation}
While previous SDS-based methods have allowed for the flexible, high-quality generation of 3D objects even with complicated prompts, they still tend to produce implausible 3D geometry. Recent studies have mitigated this issue by training multi/novel-view generative models~\cite{mvdream,zero123} on existing 3D datasets. Although these methods present viable solutions, the quality and size of existing 3D dataset is inferior in comparison to 2D data, hampering and confining the fidelity and diversity of the models directly trained on these data. This effect can be universally noted in methods that have taken the training-based approach, such as MVDream and Zero123, in which the textures of generated scenes and novel views largely retain clay-like cartoonish styles similar to that of low-quality 3D assets.


To address such issues, we explore a novel retrieval-augmented approach tailored for SDS-based frameworks, which enables the generation of high-quality 3D objects. The fundamental insight is that retrieved 3D assets, which are semantically similar to the specified text, can serve as concise references for abstract 3D appearances and geometries.

\vspace{-5px}
\subsection{Formulation}
We begin by adopting a particle-based variational inference (ParVI) framework~\cite{chen2018unified, liua2022geometry, liu2016stein, dong2022particle}, following the convention of \cite{wang2023prolificdreamer}. Within this framework, a variational distribution $\gamma$ is composed of particles $\{\theta^{(i)}\}_{i=1}^K$. Each particle is optimized using the gradient of VSD distilled from 2D diffusion models, as described in Eq.~\ref{eq:sds_ode}: $v_\mathrm{2D}^{(i)} := \nabla_{\theta^{(i)}}\mathcal{L}_\mathrm{VSD}$. Here, $v_\mathrm{2D}^{(i)}$ denotes the per-particle velocity derived from the 2D prior of the diffusion model.

Our primary goal is to enable particles to absorb meaningful information from retrieved assets $\{\theta^{(n)}_\mathrm{ret}\}_{n=1}^N$, which are conditioned on a text prompt $c$ from the 3D database $\mathcal{D}$, using the retrieval module $\xi_{N}(c,\mathcal{D})$. To achieve this, we propose a novel method to impose the velocity of each particle with the retrieved assets, as detailed in Sec.~\ref{sec:init}. This approach facilitates the subsequent optimization of the distribution $\gamma$ using the gradient derived from a lightweight-adapted 2D diffusion model, described in Sec.~\ref{sec:lightweight}. Our overall framework is illustrated in Fig.~\ref{fig:overview}.


\subsection{Initialized distribution as a geometric prior}
\label{sec:init}
Recall that the variational distribution $\gamma$ is optimized by updating the particles $\theta \sim \gamma(\theta|c)$ as in Eq.~\ref{eq:sds_ode}, and the parametrization of the score of $q^\gamma$ is additionally tuned with the particles. In this perspective, initializing the particles can be interpreted as providing a \textit{guide} for the variational distribution $q^\gamma$.

We find that retrieved neighbors can effectively act as a \textit{guide} for the variational distribution $\gamma$,  since the ideally selected nearest assets exhibit robust geometry as well as sharing semantic similarity with the optimizing particles. This approach effectively enables our model to achieve geometric robustness while overcoming the weaknesses of methods involving direct training on 3D data such as low-quality data and computation cost described above. To this end, we derive and leverage an auxiliary objective from our retrieval-augmented objective that makes $\gamma(\theta|c)$ and the empirical distribution of retrieved assets similar. The full derivation is shown in Appendix~\ref{sec:theory}.
Practically, we impose an additional velocity on each particle to coarsely initialize them during the warm-up phase as follows:
\begin{equation}
v_{\mathrm{asset}}^{(i)} := \nabla_{\theta^{(n)}}\frac{\mathbbm{1}(s\leq\tau)}{\sigma^2} \mathbb{E}_\psi \Big[ \big\|g(\theta^{(i)},\psi)- g(\theta_{\mathrm{ret}}^{(a_i)},\psi)\big\|^2_2\Big],
\label{eq:init_velocity} 
\end{equation}
where $s$, $\tau$, and $\sigma$ denote the index of iterations, the threshold for the warm-up phase, and the scaling factor, respectively. $a_n$ is a mapping function relating the $i$-th particle to its corresponding retrieved asset. Note that the particle initialization is reflected in the distribution $\gamma$ within the framework of VSD, which has the following additional objective:
\begin{equation}
\begin{aligned}
\min_{\zeta} \sum_{i=1}^{N} \mathbb{E}_{t, \epsilon, \psi} \left\| \epsilon_{\zeta,\phi} \left(x_t, t, c, \psi \right) - \epsilon \right\|_2^2,
\end{aligned}
\label{eq:vsd-lora}
\end{equation}
where $x = g(\theta^{(i)}, \psi)$. Recalling that this denoising score matching objective leads to the following relationship between $q^\gamma$ and $\epsilon_{\phi,\zeta}$: $\nabla_{x_t}q^\gamma\big(x_t|c,x=g(\theta,\psi)\big) \approx -\epsilon_{\zeta,\phi}(x_t|c,t,\psi)/{\sigma_t}$, the process in Eq.~\ref{eq:vsd-lora} can be viewed as aligning the distribution $\gamma$ with the empirical distribution $p_\xi(\theta|c)$.

The effectiveness of our initialization approach is clearly observable in Fig.~\ref{fig:nearest}, where we see that the robust geometry of the nearest assets is efficiently leveraged to ensure the robustness and consistency of corresponding particle's 3D structure. We observe that the particle's geometry and texture is not strictly confined to the initialization, allowing for freedom to make sufficient adjustments that enable the particle to faithfully follow the text prompt. 

\subsection{Lightweight adaptation of 2D prior}
\label{sec:lightweight}
During the score distillation process, viewpoint-related bias from diffusion models hinders consistent generation, but at the other extreme, fully fine-tuning diffusion models on 3D assets causes the model to lose its expressiveness. Here, we address the dilemma with \textit{lightweight} adaptation, which mostly maintains the original manifold of pre-trained diffusion models while reducing view-related biases.

A major issue that significantly hampers score distillation based methods is the fact 2D prior models are biased toward certain viewpoints, as shown in Fig.~\ref{fig:adaptation}(b), leading to text-guided predictions that are misaligned with the initial scenes. The issue of view bias of 2D prior models has been known as one of the cause of Janus problem and has been addressed in other works~\cite{armandpour2023re, hong2023debiasing}. For instance, Perp-Neg~\cite{armandpour2023re} and Debiased-SDS~\cite{hong2023debiasing} addressed this by mainly adopting negative prompt, or removing contradictory words with view prefix, respectively.

Contrary to these works, our method fortunately begins from an advantageous position, as we have access to dense renderings of 3D assets that are semantically close collected with the retrieval module. It allows us to address this issue simply yet effectively. To this end, we introduce a lightweight strategy that adapts 2D prior models by utilizing retrieved 3D assets in test time. This helps balance the probability densities across all viewpoints without a significant drop in the quality of the original 2D prior models, in despite of its simplicity.

\begin{figure}[t]
\center
\includegraphics[width=1\linewidth]{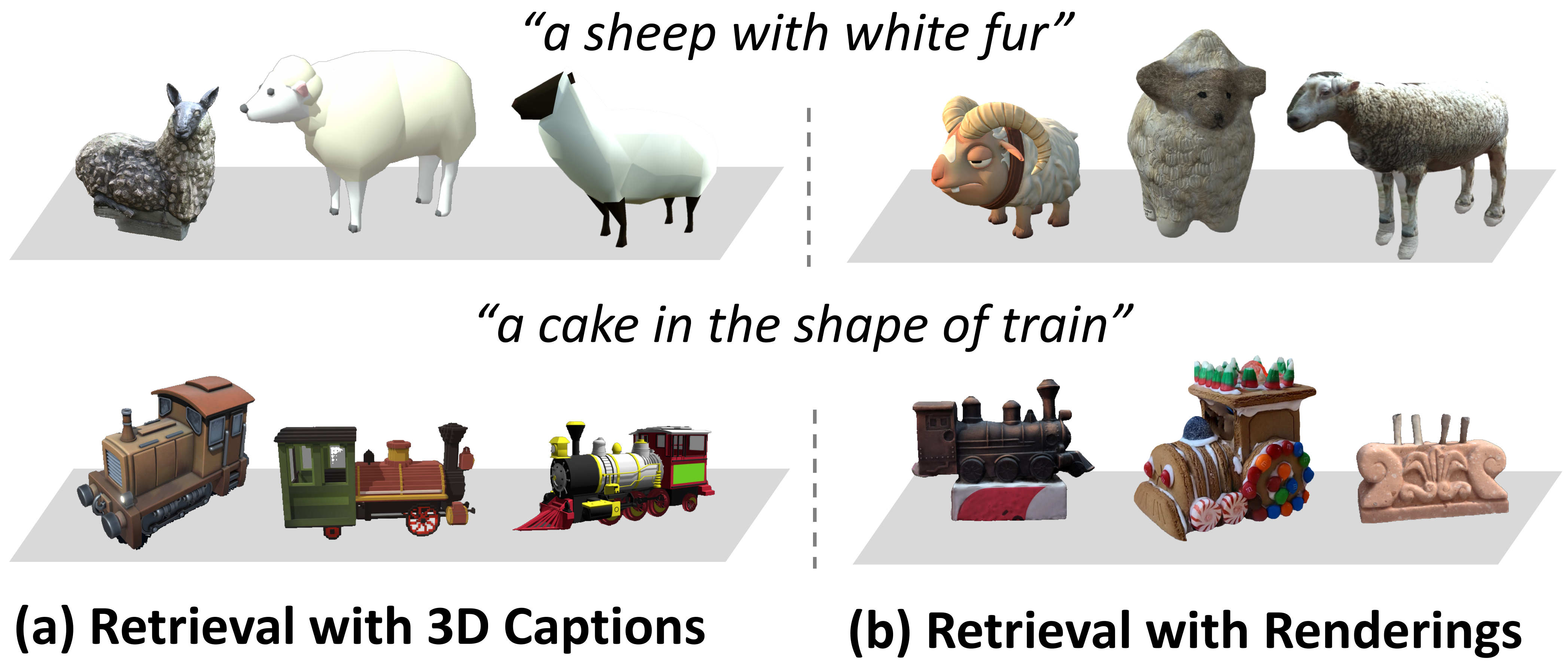}
\vspace{-20pt}
\caption{\textbf{3D Dataset retrieval.} (a) and (b) show retrieved top-$K$ nearest neighbors on CLIP-text embedding space and CLIP-image embedding space, respectively.}
\vspace{-16pt}

\label{fig:retrieval}
\end{figure}

\begin{figure*}[t]
\centering
\includegraphics[width=1.0\linewidth]{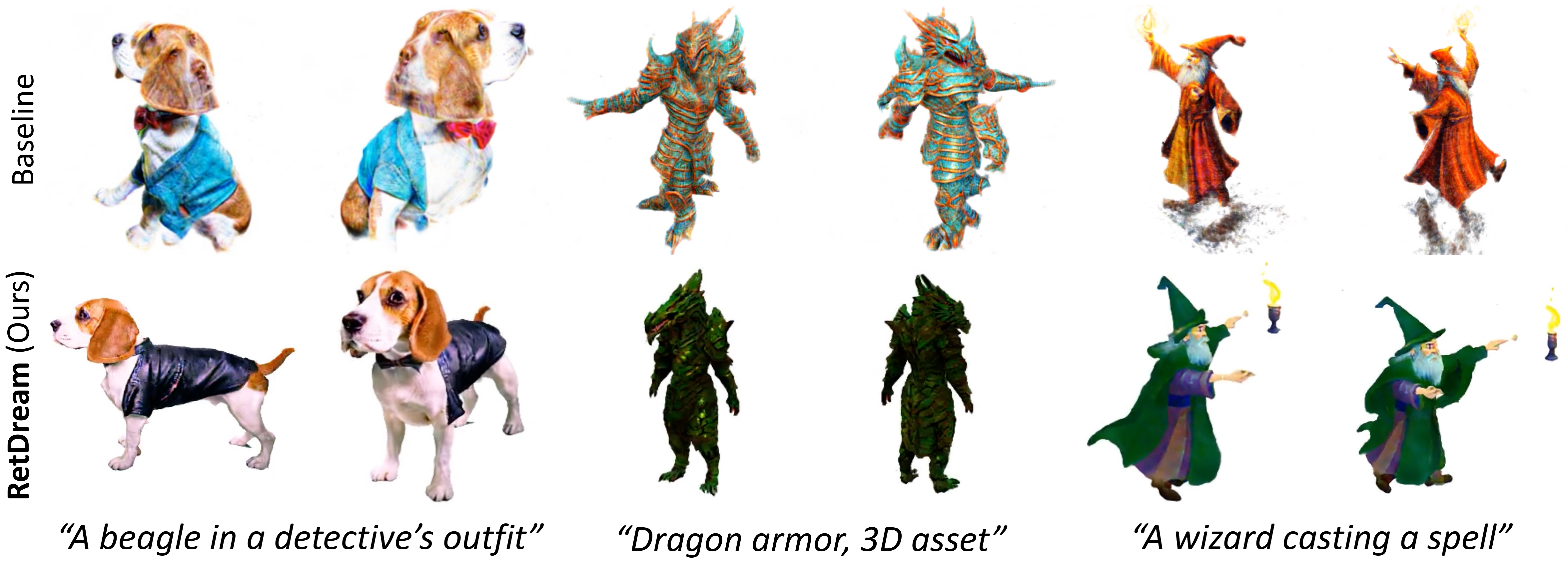}
\vspace{-10px}
\caption{\textbf{Improved 3D consistency from baseline~\cite{wang2023prolificdreamer}.} We validate the effectiveness of our approach by comparing the baseline. Given challenging prompts that are easy to cause geometric breakdowns, our results show \textbf{enhanced performance in terms of 3D}. See Project Page for videos of these results.  }
\vspace{-10pt}
\label{fig:comp_with_baseline}
\end{figure*}

Specifically, we denote $c_\mathrm{ret}^{(n)}$ as a ground-truth text caption tokens corresponding to the $n$-th retrieved asset, and $\{e_\psi\}$ as tokens of view prefixes such as ``\textit{front view}''. To obtain a adapted 2D prior \( \epsilon_{\omega,\phi} \), we densely render the retrieved assets under a uniform camera distribution, and optimize a low-rank adapter~\citep{ryu2023low,hu2021lora} with the rendered images:
\begin{equation}
\vspace{-2pt}
\begin{aligned}
\min_{\omega} \sum_{n=1}^{N} \mathbb{E}_{t, \epsilon, \psi} \left\| \epsilon_{\omega,\phi} \left(x_t, t, \mathrm{cat}(e_\psi, c_\mathrm{ret}^{(n)}) \right) - \epsilon \right\|_2^2,
\end{aligned}
\label{eq:lora}
\vspace{-3pt}
\end{equation}
where $x=g(\theta^{(n)}_\mathrm{ret}, \psi)$, and $\omega$ is a set of parameters of learnable layers inserted to the diffusion U-net. $\mathrm{cat}(\cdot)$ refers to concatenation function. At the same time, we can additionally optimize the tokens of view prefixes $\{e_\psi\}$ as well as $\omega$ using Eq.~\ref{eq:lora}. We empirically find it eliminate the model's viewpoint bias more effectively in the few-shot setting.

After the adaptation, the 2D prior $p_\phi$ used in $v_\mathrm{2D}$ is replaced with the adapted prior $p_{\omega,\phi}$ along with the learned view prefixes $\{e_\psi\}$. Our strategy demonstrates encouraging effectiveness as it shows the chronic issue of viewpoint bias in 2D prior models can be efficiently addressed thanks to the nearest neighbors without any complex technique. As shown in Fig.~\ref{fig:adaptation}, we can see samples from the adapted 2D prior is capable of generating viewpoints that more closely reflect each given view conditions without severely sacrificing its generation capability.

\subsection{Retrieval of 3D assets}
We utilize 3D assets from Objaverse 1.0~\citep{deitke2023objaverse} dataset and corresponding captions with the help of Cap3D~\cite{luo2023scalable}. We use ScaNN~\cite{avq_2020} to retrieve $N$ nearest neighbors based on CLIP embeddings~\citep{radford2021learning} of the captions and the rendered images. The query embedding can be acquired from the prompt $c$. Specifically, we utilize both image and text embeddings by performing Top-K operation with image embeddings after retrieving $N'(N'>N)$ objects with text embeddings, followed by alignment of orientations as a pre-processing step, described in detail at Appendix~\ref{sec:impl_details}. 

As we construct a list mapping UIDs of 3D assets to the corresponding CLIP embeddings, end-users can only download the retrieved 3D assets during inference, or download the whole 3D data in advance. The total time spent by the retrieval is under 3 seconds. As shown in Fig.~\ref{fig:nearest} and Fig.~\ref{fig:retrieval}, in situations where completely matching 3D assets are not retrieved given challenging prompts, we found it still shows sufficient performance to serve as references for generation.
\vspace{-10pt}

\begin{figure}[t]
\centering
\includegraphics[width=\linewidth]{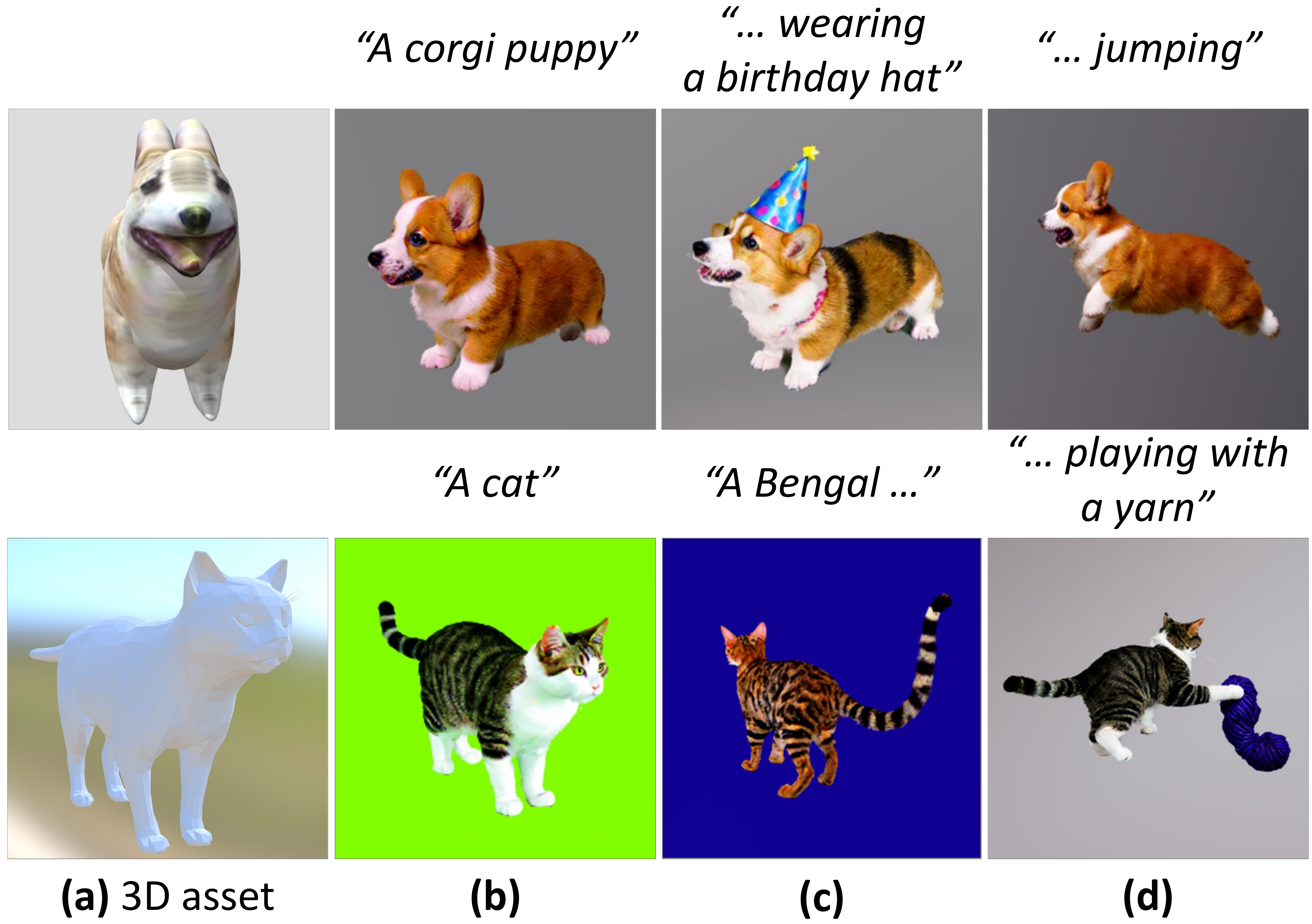}
\vspace{-15pt}
\caption{\textbf{Variations of text prompts with fixed 3D asset.} Given the retrieved 3D asset shown at leftmost column, each column represents separate optimization results given different text conditions. Note that the scene under optimization is not strictly constrained to the asset, but retains strong capability to generate a 3D scene relevant to the given text prompt and the assets.}
\label{fig:variations}
\vspace{-15pt}
\end{figure}

\begin{figure}[t]
\centering
\vspace{-5px}
\includegraphics[width=1\linewidth]{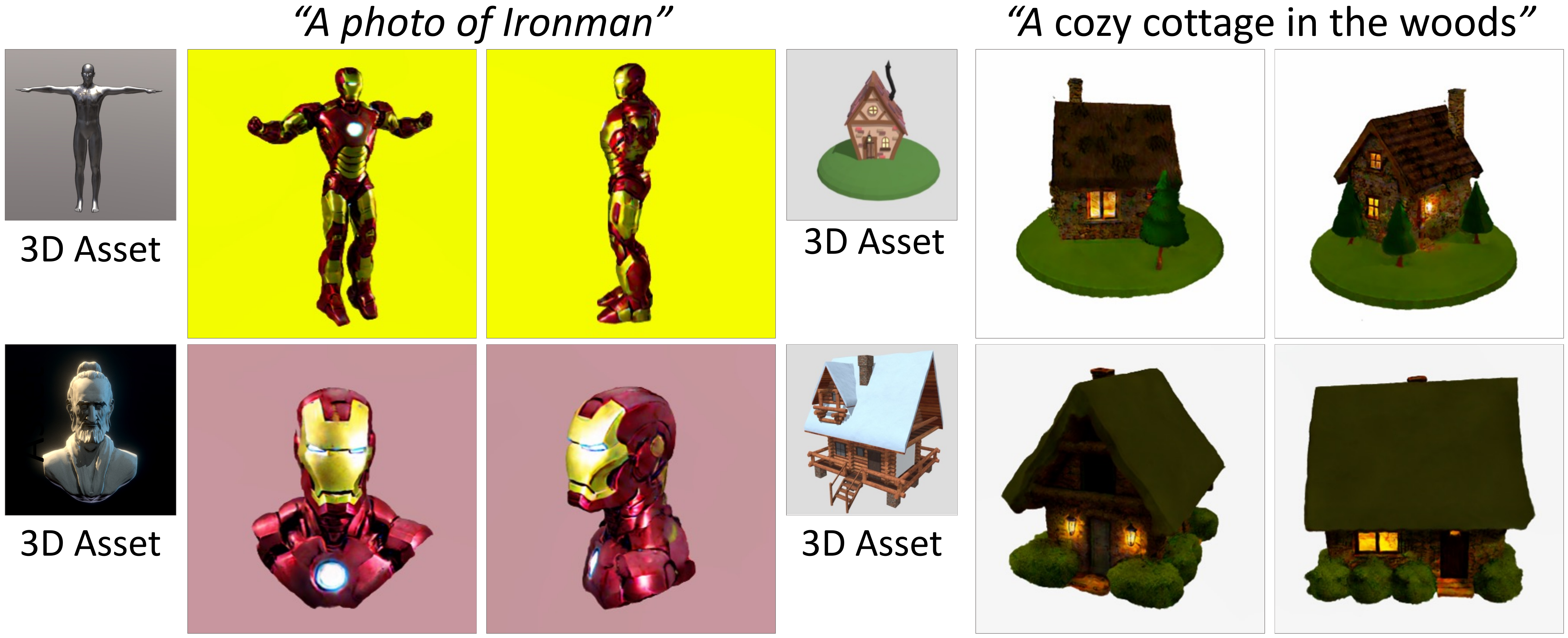}
\vspace{-15pt}
\caption{\textbf{Variations of 3D assets with fixed text prompt.} We fix the prompts and vary the assets that correspond to each particle. This shows how our method is effected by the retrieved assets.}
\label{fig:change}
\vspace{-5px}
\end{figure}

\vspace{-5px}
\section{Analysis}
In this section, we provide extensive analyses on the properties of our approach, including qualitative and quantitative evaluations. The implementation details are described in Appendix~\ref{sec:exp_setup} and \ref{sec:impl_details}.

\begin{figure*}[t]
\centering
\includegraphics[width=0.95\linewidth]{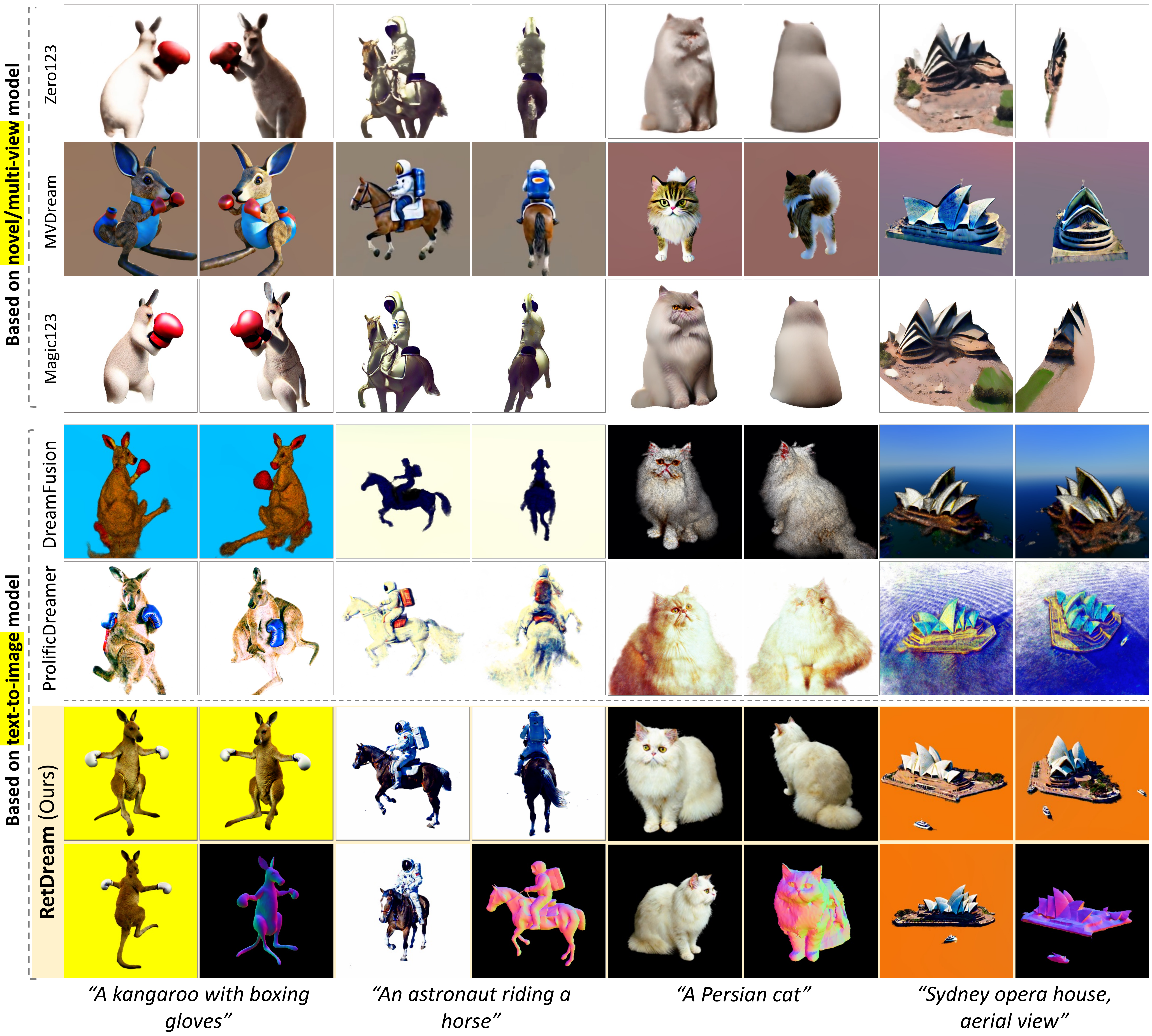}
\vspace{-12pt}
\caption{\textbf{Comparison with other works.} We compare our framework with novel/multi-view model based frameworks: Zero123~\cite{zero123}, MVDream~\cite{mvdream}, Magic123~\cite{qian2023magic123}, and image generative model based frameworks: DreamFusion~\cite{poole2022dreamfusion}, ProlificDreamer~\cite{wang2023prolificdreamer}. \textbf{When zoomed in, the 3D inconsistency and resulting artifacts are most noticeable}. We provide accompanying \textbf{video results} in Project Page. Also, more qualitative results can be found in Appendix~\ref{sec:addqual_app}.}
\vspace{-12pt}
\label{fig:main_comp}
\end{figure*}

\vspace{-10pt}
\paragraph{Does it handle corner cases of the baseline?}
One of our goals is to alleviate 3D inconsistency, which frequently occurs when given challenging prompts. For example, when testing both the baseline~\cite{wang2023prolificdreamer} and our method on generating creatures with a face, we observe that our approach can generate more plausible outputs, as illustrated in Fig.~\ref{fig:comp_with_baseline}. Consistent with our claims in Sec.~\ref{sec:method}, we corroborate that our method alleviates such issues by utilizing a retrieval-integrated prior.

\vspace{-10pt}
\paragraph{Influence from retrieved assets.} 
In our retrieval-based approach, we address two critical questions: whether there's an over-reliance on retrieved assets leading to overly constrained results, or whether these assets fail to remarkably influence the outcome. To explore the impact of retrieved assets, we set the number of assets for retrieval $N$ to $1$, and gradually change the corresponding text prompt to be distanced from the asset. Fig.~\ref{fig:variations} clearly shows our observation; our approach flexibly operates depending on the similarity between the text prompt and the retrieved asset. In (b) of Fig.~\ref{fig:variations}, where the text prompt aligns best with the asset, we observe minimal geometric changes and textural variations, whereas in (c) and (d), sufficient adjustments are made where necessary. Additionally, Fig.~\ref{fig:change} shows the results by changing the assets with the fixed prompts. It also supports our observation, showing the flexibility of our approach.

\begin{figure}[t]
\centering
\vspace{-5px}
\includegraphics[width=1\linewidth]{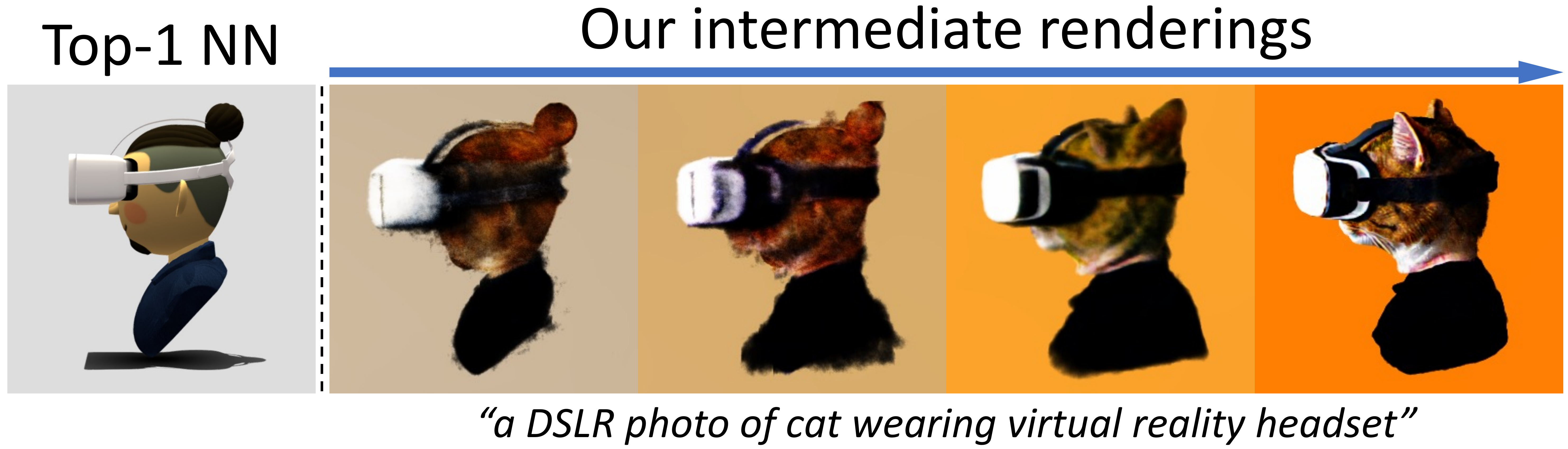}
\vspace{-15pt}
\caption{\textbf{Intermediate renderings in optimization.} We visualize the intermediate renderings of the particle which corresponds to top-1 retrieved asset. Geometric influence of the nearest assets is significant when the 3D representation is coarse, and fine details are generated through the adapted 2D prior. Details are clearest when zoomed in.}
\label{fig:con_vis}
\vspace{-10pt}
\end{figure}

\vspace{-10pt}
\paragraph{Qualitative evaluation.}
We compare our methods with state-of-the-art text-to-3D~\cite{wang2023prolificdreamer, poole2022dreamfusion} and image-to-3D methods~\cite{zero123, mvdream, qian2023magic123}. In the case of image-to-3D methods, we carefully selected appropriate images generated by the text-to-image model~\cite{rombach2022high}. These generated images are delineated in Fig.~\ref{fig:imageto3d}. Comparative results are shown in Fig.~\ref{fig:main_comp}. In contrast to preceding text-to-image prior based methods, our framework shows enhanced geometric consistency. On the other hand, while methods employing novel/multi-view models yield plausible geometry, they often suffer from degraded texture, such as overly smoothed surfaces, which detracts from realism, whereas ours generates high-quality textures. Additioanlly, we visualize the optimization process by showing the intermediate renderings of the particle with the corresponding 3D asset in Fig.~\ref{fig:con_vis}.

\paragraph{Quantitative evaluation.}
Currently, there is no established metric for evaluating the open-domain text-to-3D field, as text-to-3D is an inherently subjective task and encompasses various aspects that are challenging to quantify. Nevertheless, we align with the practices of quantitative evaluation~\cite{poole2022dreamfusion, li2023instant3d, yu2023text} in text-to-3D works by utilizing CLIP-based metrics for our quantitative assessments. Specifically, we measure the average CLIP score between text and 3D renderings using variants of the CLIP model,   OpenCLIP ViT-L/14 trained on DataComp-1B~\cite{ilharco_gabriel_2021_5143773} and CLIP ViT-L/14~\cite{radford2021learning}. The evaluation is done with 50 prompts, each rendered with 120 viewpoints of the corresponding 3D outputs. We note that the CLIP model for retrieval is not used for the evaluation. For view consistency, some works~\cite{li2023sweetdreamer} manually check their success rate, and \cite{hong2023debiasing} proposes A-LPIPS, an average LPIPS~\cite{zhang2018perceptual} between adjacent images of generated 3D scenes to measure artifacts caused by view inconsistency. We adopt A-LPIPS as an alternative metric to quantify view consistency and report it alongside the CLIP score in Tab.~\ref{table:quan}, showing ReDream exhibits superior performance in terms of text-3D alignment and view consistency.

\begin{table}[t]
\small
\centering
\begin{tabular}{lcccc} 
\toprule
\multirow{2}{*}{Methods} & \multicolumn{2}{c}{CLIP-Score $\uparrow$} & \multicolumn{2}{c}{A-LPIPS $\downarrow$} \\
 & \scriptsize CLIP L/14 & \scriptsize OpenCLIP L/14 & \scriptsize VGG &   \scriptsize Alex \\
\midrule
DreamFusion & 0.242 & 0.185 & 0.075 & 0.076       \\
MVDream & 0.263 & 0.217 & 0.062 & 0.072      \\
ProlificDreamer & 0.218 & 0.204 & 0.227 & 0.135       \\
\textbf{ReDream} & \textbf{0.274} & \textbf{0.227}& \textbf{0.041}& \textbf{0.054} \\
\bottomrule
\end{tabular}
\vspace{-10pt}
\caption{\textbf{Quantitative evaluation.} We compare our approach with recent text-to-3D works~\cite{poole2022dreamfusion, mvdream, wang2023prolificdreamer}. CLIP-score indicates the alignment between text and 3D, while A-LPIPS represents the degree of artifacts due to 3D inconsistency~\cite{hong2023debiasing}. 
}
\vspace{-5pt}
\label{table:quan}
\end{table} 

\begin{table}[!t]
\vspace{6pt}
\small
\centering
\begin{tabular}{lc} 
\toprule
Methods & Preference \\
\midrule
Baseline~\cite{wang2023prolificdreamer} & 24.7\%      \\
\textbf{Ours} & \textbf{75.3\%} \\
\bottomrule
\end{tabular}
\caption{\textbf{User study.} We report the percentage of user preference from 92 participants.}
\vspace{-5pt}
\label{table:userstudy}
\end{table} 



\paragraph{User study.}
We conduct a user study with 92 participants; the result is shown in Tab.~\ref{table:userstudy}. Each participant is asked seven randomly selected questions. Specifically, we inquire about their preference between our method and the baseline, taking into account geometry and textural fidelity. Approximately 75\% of the participants express a preference for the results by our method over the baseline. More details are described in Appendix~\ref{sec:user_study}.

\paragraph{Ablation on each component.}
We conduct an ablation study on each component of our pipeline, as depicted in Fig.~\ref{fig:abl} of Appendix~\ref{sec:ablation_app}. We observe that initializing the variational distribution is crucial for the overall geometry, and lightweight adaptation effectively reduces artifacts such as eyes on the back.

\paragraph{2D experiments on lightweight adaptation.}
We conduct a 2D experiment to detail the process of lightweight adaptation, which is depicted in Fig.~\ref{fig:abl_light}. We also report our analysis of how a 3D asset influences the 2D prior model in lightweight adaptation in Fig.~\ref{fig:light_overfit_test}. Specifically, we progressively change the prompts to describe other objects with different textures while keeping the used asset constant. The results suggest that the adaptation primarily concentrates on general aspects, such as viewpoint, instead of focusing on particular details like texture. The details are described in Appendix~\ref{sec:ablation_light_app} and \ref{sec:light_app}, respectively.

\vspace{-5pt}
\section{Conclusion}
We present a novel retrieval-based framework for text-to-3D generation in which retrieved assets are used as efficient guidance for enhanced fidelity and geometric consistency of generated 3D scenes. We propose simple, elegant methods to leverage the retrieved assets for aforementioned purpose, which use the retrieved asset as initializing point of 3D scene's variational distribution, and also use it for adaptation of 2D diffusion model toward increased faithfulness to given view prompts for enhanced view consistency and reduction of the Janus problem. Our approach does not require extensive fine-tuning or compromising of the capabilities of 2D diffusion models, offering a promising avenue for future developments in this domain. Through extensive experiments and analysis, both quantitative and qualitative, we demonstrate that our model successfully achieves the goal of quality improvement and geometric robustness in text-to-3D generation.

\section*{Impact Statements}
This paper presents in the field of AIGC (AI-generated Content) aiming for research advancements. While there may be potential social impacts as a consequence, there is nothing in particular to be highlighted. The framework presented in this paper utilizes data retrieved from an external database; therefore, users employing this framework must verify the copyright of the database they use.

\bibliography{main}
\bibliographystyle{icml2024}

\newpage
\appendix
\onecolumn
\section{Experimental Setup}
\label{sec:exp_setup}
We build our method upon ProlificDreamer~\citep{wang2023prolificdreamer}, and follow \citep{threestudio2023} for details of the implementation. Our experiments were conducted on an NVIDIA RTX A6000 GPU, with a total of 20,000 iterations of optimization for generation. For all our experiments, Instant-NGP~\citep{muller2022instant} is used for our NeRF backbone and Stable Diffusion v2~\citep{rombach2022high} as the 2D prior. For our method, we retrieve 3 assets and render our retrieved data with 100 uniformly sampled camera poses. We compare our framework with various methods~\cite{poole2022dreamfusion, wang2023prolificdreamer, zero123, mvdream, qian2023magic123}. For \cite{wang2023prolificdreamer, zero123, mvdream, qian2023magic123}, we utilize author-provided implementations. For \cite{poole2022dreamfusion}, we used Stable Diffusion  as a 2D prior model on Threestudio~\cite{threestudio2023}, as Imagen~\cite{saharia2022photorealistic} used in their implementation is not publicly available.

\section{Additional Implementation Details}
\label{sec:impl_details}
\paragraph{3D retrieval procedure.}
3D datasets such as Objavarse~\cite{deitke2023objaverse} are too large for users to download; therefore, we construct a list that includes the UIDs of the 3D contents along with the corresponding CLIP embeddings of the renderings and captions~\cite{luo2023scalable}, and we proceed with retrieval employing the ScaNN~\cite{avq_2020} algorithm. In this case, the total time spent by the retrieval is under 3 seconds, which is negligible compared to the time taken for the entire generation process. During inference, we download and load only the essential 3D assets using the UIDs acquired via the retrieval process.

While the Objaverse dataset offers a variety of 3D assets, their orientations are generally not aligned. We observe that this is not necessarily an issue, since score distillation works with misaligned orientations as well. Nevertheless, before employing our nearest neighbors, we find it beneficial to align their frontal views. Specifically, we can categorize 3D objects into (1) those where the front is distinguishable, such as objects with a clear frontal aspect, and (2) those where the front is not distinguishable, such as radially symmetric objects. In the case of the latter, the importance of the view prefix is not high. This is because it is not only difficult to semantically predict the front views but also because the necessity to find their orientations is not significant, allowing it to be disregarded. For the former case, it is relatively important to identify the semantic fronts to assign appropriate view prefixes to their renderings.

For this purpose, we compute the CLIP similarity score between the prompts with view prefixes ``\textit{front view}'', ``\textit{side view}'', ``\textit{back view}'' and the rendered images with different camera poses. Subsequently, we rotate the 3D assets according to the camera poses that exhibit the relatively highest CLIP similarity score. Despite its simplicity, this method effectively aligns our retrieved assets. Note that objects with semantically indistinct front and back differences (such as an ice cream cone) exceptionally demonstrated lower accuracy levels. Nevertheless, due to the nature of such objects, the necessity for orientation alignment is less critical, and we found that this has a minimal impact on the performance of the final results. For the performance of the alignment, refer to Sec.~\ref{sec:or_align}.

\paragraph{Additional regularization.}
Concerning the degree to which particles may deviate from or overlook their initial state, this divergence becomes apparent when the bias of the 2D prior towards a particular text prompt continues to steer away from \(v_\mathrm{asset}\). To alleviate this problem, we adopt a variant of the delta denoising score~\cite{hertz2023delta}, initially employed in image editing, in 3D cases. Specifically, \(v_\mathrm{2D}(\theta=\theta_0)\) represents the predicted velocity (gradient) of the 2D prior at the point \(\theta_0\). Ideally, the combination of a retrieved asset and text should result in minimal gradient or velocity, leading us to identify \(v_\mathrm{2D}(\theta=\theta_\mathrm{ret})\) as a noisy component. To reduce the artifacts, we adjust the original \(v_\mathrm{2D}\) by subtracting from it: $\Tilde{v}_\mathrm{2D} := v_\mathrm{2D} - v_\mathrm{2D}(\theta=\theta_\mathrm{ret})$. We opt for updates using $\Tilde{v}_\mathrm{2D}$ in place of $v_\mathrm{2D}$ for every three iterations. We found that the adjustment strength can be effectively controlled by modulating the frequency of these updates and adjusting the weight.

\section{Conceptual Analysis of Our Approaches}
\label{sec:theory}
\paragraph{Preliminary.}
Here, we formulate text-to-3D generation with score distillation~\citep{poole2022dreamfusion}, which leverages a diffusion model~\citep{saharia2022photorealistic,rombach2022high} as a prior to optimize a 3D representation for a given text.
We extend the framework of Variational Score Distillation (VSD)~\cite{wang2023prolificdreamer}, which generalizes the original Score Distillation Sampling (SDS)~\cite{poole2022dreamfusion}.

VSD aims to optimize the distribution of 3D representations given a text prompt, while SDS~\cite{poole2022dreamfusion} aims to optimize an instance of 3D representation for text-to-3D generation. 
We also define $q^\gamma(x|c,\psi)$ as an implicit distribution of the rendered image $x:=g(\theta,\psi)$ where $\theta \sim \gamma(\theta|c)$.
Then, VSD minimizes the variational objective, $D_\mathrm{KL}\big(q^\gamma(x|c) || p_\phi(x|c)\big)$ to find an optimal $\gamma^{*}$, where $q^\gamma(x|c)$ is marginalized distribution w.r.t. camera viewpoints $p(\psi)$ and $p_\phi(x|c)$ is empirical likelihood of $x$ estimated by a diffusion model $\phi$. 
Since the diffusion model learns noisy distribution $p_\phi(x_t|c,t)$ according to diffusion process~\citep{ho2020denoising,song2020score}, the variational objective can be decomposed as follows:
\begin{equation} \label{eq:var_obj}
    \gamma^*:= \underset{\gamma}{\mathrm{arg}\min}\,\mathbb{E}_t\Big[(\sigma_t/\alpha_t)w(t)D_\mathrm{KL}(q_t^\gamma(x_t|c) || p_\phi(x_t|c,t))\Big],
\end{equation}
where $q_t^\gamma(x_t|c)$ is a noisy distribution at noise level $t$ following the diffusion process. 

VSD employs the particle-based variational inference (ParVI)~\citep{chen2018unified,liua2022geometry,wang2019function,dong2022particle} to minimize Eq.~\ref{eq:var_obj}.
The minimization process proceeds via a Wasserstein gradient flow~\citep{chen2018unified}.
Specifically, $N$ particles $\{\theta^{(i)}\}_{i=1}^N$ are first sampled from initial $\gamma(\theta|c)$, and then updated with the following ODE:
\begin{equation}
    v_\mathrm{2D}:=\frac{d\theta_\eta}{d\eta}=-\mathbb{E}_{t,\epsilon,\psi}\Big[w(t)\big(-\sigma_t\nabla_{x_t}\log p_\phi(x_t|c,t) - (-\sigma_t\nabla_{x_t}\log q_t^{\gamma_\eta}(x_t|\psi,c))\frac{\partial g(\theta_\eta,\psi)}{\partial \theta_\eta}\big)\Big],
\label{eq:v_prior}
\end{equation}
where $\eta$ denotes ODE time such that $\eta\geq0$, and the distribution $\gamma_\eta$ converges to an optimal distribution $\gamma^*$ as $\eta\to\infty$ and $\theta_\eta$ is sampled from $\gamma_\eta$. 
Note that the first term is a score of noisy real image, approximated by a predicted score of the diffusion model $\epsilon_\phi(x_t,c,t)$. 
The second term can be regarded as a score of noisy rendered images. They parameterize the second term to a score-predicting U-shaped network. Practically, they train the U-Net network from the pretrained diffusion model with low-rank adaptation (LoRA), $\epsilon_{(\phi,\zeta)}(x_t,t,c,\psi)$, where $\zeta$ is a set of parameters of trainable residual layers for LoRA. VSD allows to generate realistic textures of 3D object given a text, but we remark that these method are still vulnerable to generating unrealistic geometry.

\paragraph{Regarding total velocity comprised of $v_\mathrm{asset}$ and $v_\mathrm{2D}$.}
We first show a total velocity in warm-up phase can be roughly interpreted as minimizing the distance between the variational distribution $\gamma$ and our retrieval-integrated prior we present in the followings. Let $\xi_{N}(c,\mathcal{D})$ be a non-parametric sampling strategy to obtain the $N$ nearest neighbors using the retrieval algorithm conditioned on text prompt $c$ in the 3D dataset $\mathcal{D}$. Our goal is to integrate the rich view-dependent information from the retrieved assets with that of 2D prior models, and derive the particle-based optimization process for the variational distribution $\gamma(\theta|c)$. We assume the probability density of 3D content $\theta$ by 2D prior is proportional to the expected densities of its multiview images w.r.t. camera viewpoints, following ~\cite{wang2023score}:
\begin{equation}
    p_\phi(\theta|c) \propto \mathbb{E}_{\psi}\big[ p_\phi^\mathrm{2D}(x| c, x=g(\theta, \psi) \big].
\end{equation}
\begin{wrapfigure}{r}{6.0cm}
  \centering
  \includegraphics[width=1.0\linewidth]{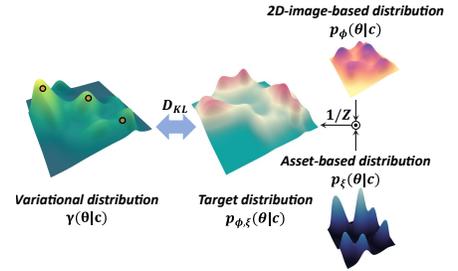}
  \caption{\textbf{Conceptual figure of the variational objective.} Geometrically plausible areas by retrieved nearest neighbors have higher density in the target distribution.}
  \label{fig:conceptual}
\end{wrapfigure}
Technically, this expectation is set as the geometric expectation (see the last paragraph of this section for details). 
Subsequently, let us consider a following energy functional for integrating the retrieved assets: 
\begin{equation}
     \mathcal{E}[\gamma] := D_\mathrm{KL}\big( \gamma(\theta|c) || p_{\phi, \xi}(\theta|c)\big),
\end{equation}
where we present $p_{\phi, \xi}(\theta|c)$ as a retrieval-integrated prior. Based on the intuition that a 3D asset selectively filters a distribution, we simply multiply and normalize the two distributions:
\begin{equation}
    p_{\phi, \xi}(\theta|c) := \frac{1}{Z'} p_\phi(\theta|c) p_\xi(\theta|c),
\end{equation}
where we denote $p_\xi(\theta|c)$ as a 3D likelihood from the retrieved assets, 
and $Z'$ denotes the normalizing constant. Fig.~\ref{fig:conceptual} depicts the intuition behind this; the distribution $p_\xi$ derived from the retrieved nearest neighbor serves as an implicit filter for plausible geometry. 

Specifically, we derive the distribution $p_\xi(\theta|c)$ from an empirical distribution defined over the top-$N$ nearest neighbors $\{\theta_\mathrm{ret}^{(n)}\}_{n=1}^{N}$ utilizing the sampling strategy $\xi_{N}(c,\mathcal{D})$, then applying non-parametric kernel $K$ for density estimation. Intuitively, the likelihood $p_\xi(\theta|c)$ depicts how close the particle is to the retrieved assets.

Using the definition of KL divergence, this is further expanded:
\begin{align}
\mathcal{E}[\gamma]&=\mathbb{E}_{\psi}[D_\mathrm{KL}\big( q^\gamma(x|c) ||  p_\phi^\textrm{2D}(x|c)\big)] + H\big(\gamma(\theta|c);p_\xi(\theta|c)\big) - C
\\&= \mathbb{E}_{\psi}[D_\mathrm{KL}\big( q^\gamma(x|c) ||  p_\phi^\textrm{2D}(x|c)\big)] - \mathbb{E}_{\gamma(\theta|c)} \left[ \log\sum_n K (\theta-\theta_\mathrm{ret}^{(n)}) \right] - C',
\label{eq:expand}
\end{align}
where $x=g(\theta,\psi)$, and $H$ is the joint entropy. 
$C$ and $C'$ are constants to be unnecessary.

The minimization process then proceeds via a Wasserstein gradient flow~\citep{chen2018unified}. 
Given $\mathcal{E}[\gamma_\eta]$ at an optimization step $\eta$, the velocity of particles, $v_\eta := \frac{d\theta_\eta}{d\eta} = \nabla_\theta \frac{\delta \mathcal{E}[\gamma_\eta]}{\delta \gamma_\eta}$, is obtained by calculating the functional derivative $\frac{\delta \mathcal{E}[\gamma_\eta]}{\delta \gamma_\eta}$ as follows:
\begin{align}
{v}_\eta &= \nabla_\theta \frac{\delta \mathcal{E}[\gamma_\eta]}{\delta \gamma_\eta} = v_{\textrm{2D}} - \nabla_\theta \log\sum_n K (\theta-\theta_\mathrm{ret}^{(n)}) \\&= v_{\textrm{2D}} + {v}_{\textrm{asset}}.
\label{eq:velocity-asset}
\end{align}
where $v_\mathrm{asset}$ is the velocity derived from retrieval, and $v_\mathrm{2D}$ is derived as in Eq.~\ref{eq:v_prior}.  $K(\cdot)$ can be any kernel function. This suggests that the total velocity of particles, derived from the variational objective with the augmented distribution, actually consists of the two components.

As one usual choice would be Gaussian kernel, we start by choosing $K$ as Gaussian kernel with a variance $\sigma^2$. However, in practice, strictly computing the derived {$v_\mathrm{asset}$ with Gaussian kernel for all assets remains inefficient, given that it is defined in a high-dimensional space. To address this inefficiency, we turn to our observation that the direction of the velocity of each particle is largely determined by its random initialization as it is drawn towards the nearest mode, which suggest a feasible alternative. Motivated by this observation, instead of computing all terms, we use an efficient surrogate method to compute $ v_{\mathrm{asset}} $ for each particle as follows:
\begin{equation}
v_{\mathrm{asset}}^{(i)} = \sum_{n}\frac{\pi^{(i)}_{n}}{\sigma^2}(\theta^{(i)}-\theta_{\mathrm{ret}}^{(n)}) = \frac{1}{\sigma^2}\sum_{n}\pi^{(i)}_{n}(\theta^{(i)}-\theta_{\mathrm{ret}}^{(n)}),
\label{eq:velocity} 
\end{equation}
where $\theta^{(i)}$ is $i$-th particle from the variational distribution $\gamma(\theta|c)$ and we assign to them one-hot vectors \( \pi \) whose non-zero indices correspond to a closest random asset when initialized.
Intuitively, this property of a particle to follow a specific mode is determined at the time of its creation.

For generality, the particle $\theta^{(i)}$ and 3D asset $\theta_{\mathrm{ret}}^{(n)}$ have not been assumed to have specific representations (\textit{e.g.}, NeRF~\citep{mildenhall2021nerf}, DMTet~\citep{dmtet}, or mesh), and could be different representations. However, some representations can be only partially observed through the differentiable rendering function $g$. Accordingly, in Eq.~\ref{eq:velocity}, the shift term is given in the form of a gradient with respect to the objective~\citep{tancik2021learned}:
\begin{align}
(\theta^{(i)}-\theta_{\mathrm{ret}}^{(n)}) &\simeq \nabla_{\theta^{(i)}} \mathbb{E}_\psi \Big[ \big\|g(\theta^{(i)},\psi)- g(\theta_{\mathrm{ret}}^{(n)},\psi)\big\|^2_2\Big].
\label{eq:partial}
\end{align}
Consequently, the velocity of $i$-th particle towards the retrieved asset in warm-up phase becomes to:
\begin{equation}
v_{\mathrm{asset}}^{(i)} \simeq \nabla_{\theta^{(n)}} \frac{1}{\sigma^2}  \sum_{n}\pi^{(i)}_{n}  \mathbb{E}_\psi \Big[ \big\|g(\theta^{(i)},\psi)- g(\theta_{\mathrm{ret}}^{(a_i)},\psi)\big\|^2_2\Big].
\end{equation}

\paragraph{Lightweight adaptation as a parametric approach.}
In the Lightweight adaptation introduced in Sec. 4.4 of the main paper, the adaptor of the 2D prior can be interpreted as a parametric model moderately reflecting $p_\xi(\theta|c)$. Specifically, given the original relationship of the pretrained diffusion models~\cite{song2020score,ho2020denoising},
\begin{equation}
\nabla_{x_t} \big[ p_{\phi}(x_t|c,\psi) \big] \approx -\frac{\epsilon_{\phi}(x_t,t,c)}{\sigma_t},
\end{equation}
and given the (variational) objective of the lightweight adaptation,
\begin{equation}
\begin{aligned}
\sum_{n=1}^{N} \mathbb{E}_{t, \epsilon, \psi} \left\| \epsilon_{\omega,\phi} \left(x_t, t, \mathrm{cat}(e_\psi, c_\mathrm{ret}^{(n)}) \right) - \epsilon \right\|_2^2,
\end{aligned}
\label{eq:lora_appendix}
\end{equation}
where $\epsilon_{\omega,\phi}$ represents LoRA, whose initialization is exactly the same function as $\epsilon_{\omega}$. Since the model $\epsilon_{\omega,\phi}$ implicitly matches the empirical distribution $p_{\xi}$ of retrieved assets, and because we early-stopped the training to maintain quality, $p_{\xi}$ is moderately reflected. In other words, the score (inclination) of $p_\xi$ is moderately learned by the adapted diffusion model.

Consequently, the resulting velocity from the adapted 2D model can be derived in the same way as in~\cite{wang2023prolificdreamer}:
\begin{equation}
\begin{aligned}
\hat{v}_\mathrm{2D}^{(n)} &:= \nabla_{\theta^{(n)}} \frac{\delta}{\delta\gamma} \mathbb{E}_{t,\epsilon,\psi}\Big[ D_\mathrm{KL}\big(q^\gamma(x_t|c) || p_{\xi,\omega}(x_t|c,\psi)\big)\Big]
\\ &= -\mathbb{E}_{t,\epsilon,\psi}\Big[w(t)\big(  \epsilon_{\omega,\phi} (x_t, t, \mathrm{cat}(e_\psi, c) ) - \epsilon_{\phi,\zeta}(x_t|c,t,\psi)\big)\frac{\partial g(\theta,\psi)}{\partial \theta}\Big],
\end{aligned}
\end{equation}
With this in mind, as our the previous approach, the velocity attributable to 3D assets can be separated:
\begin{equation}
\begin{aligned}
\hat{v}_\mathrm{3D}^{(n)} &:= \hat{v}_\mathrm{2D}^{(n)} - {v}_\mathrm{2D}^{(n)}.
\end{aligned}
\end{equation}

\paragraph{Assumption on the density function of 3D content.}
Several works~\cite{wang2023score,hong2023debiasing} have clarified the assumptions on the density function of 3D content, which is an important part in lifting the 2D generative models to do 3D generation. Specifically, SJC~\cite{wang2023score} proposes to assume it to be proportional to an arithmetic expectation of likelihoods over camera points, \textit{i.e.}, $p_\phi(\theta|c) \propto \mathbb{E}_\psi [p_\phi^\textrm{2D}(x|c,x=g(\theta,\psi))]$, and D-SDS~\cite{hong2023debiasing} finds it more beneficial to define it as a product of likelihoods over a set of camera points. In this paper, we instead use the geometric expectation. Actually, all three premises do not affect the solution of the minimization or maximization problem of the logarithm. Besides, in terms of KL divergence, setting the target distribution to the geometric mean has the following benign property:
\begin{equation}
D_{\text{KL}}(q||\kappa \mathbb{G}_\psi [p_\phi^\textrm{2D}(x|c,x=g(\theta,\psi))) = \mathbb{E}_\psi[D_{\text{KL}}(q||p_\phi^\textrm{2D}(x|c,x=g(\theta,\psi)))] - \log \kappa,
\end{equation}
where $\kappa$ is a constant.
\section{Additional Experimental Details}
\label{sec:exp_details}

\paragraph{Reference images for image-to-3D works~\cite{zero123,qian2023magic123} in Fig.~\ref{fig:main_comp}.}
In the domain of SDS-based task, some works~\cite{zero123,qian2023magic123} that address 3D consistency essentially receive images as inputs for image-to-3D, which complicates direct comparisons with text-to-3D works. However, as mentioned in Zero123~\cite{zero123}, it is possible to indirectly facilitate text-to-3D by first generating images from text using text-to-image generation models like Stable Diffusion~\cite{ldm}. In this context, we adopt such an approach in Fig. 7 of our main paper, providing a qualitative comparison with Zero123~\cite{zero123} and Magic123~\cite{qian2023magic123}. For the sake of fairness, we disclose the reference images generated by Stable Diffusion in Fig.~\ref{fig:imageto3d}. These images have undergone processing such as background removal, in accordance with the method described in Zero123~\cite{zero123}.

\begin{figure}[h]
\centering
\includegraphics[width=0.8\linewidth]{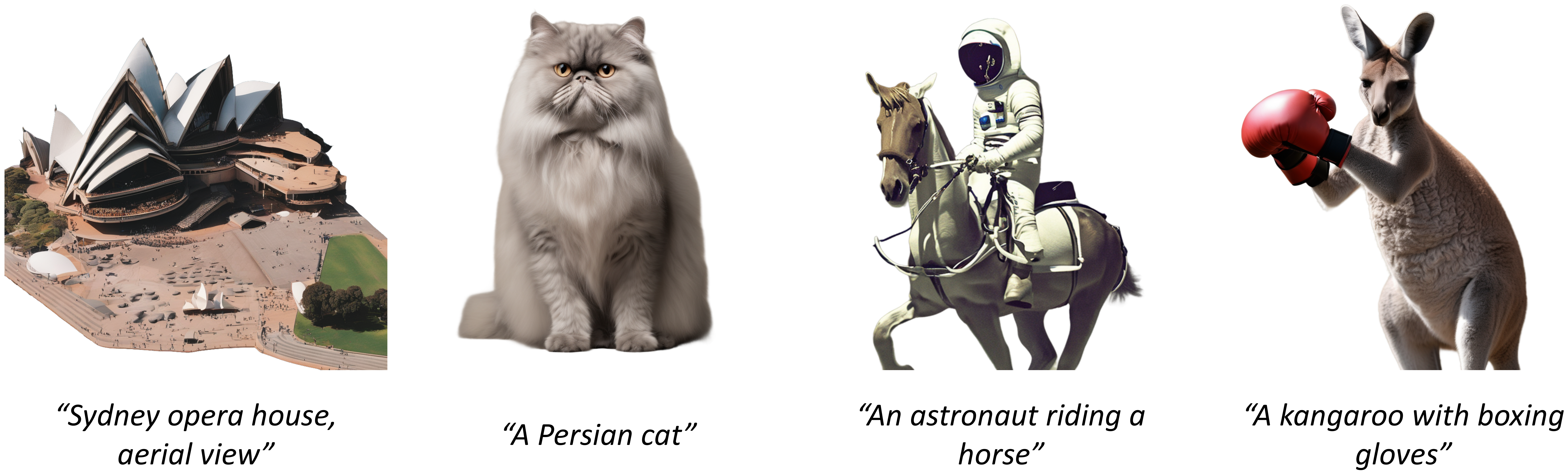}
\caption{\textbf{Reference images for image-to-3D works~\cite{zero123,qian2023magic123}.} These images are generated by Stable Diffusion~\cite{ldm}, followed by processing such as background removal. }
\label{fig:imageto3d}
\end{figure}

In certain cases, we have observed results of image-to-3D methods that do not reach the quality of the qualitative results shown in their paper. We conjecture that this is due to the sensitivity of the input images generated from text-to-image models when they diverge from the domain of the training (or fine-tuning) dataset. In contrast, the results of text-to-3D tasks, including our results, seem not to encounter this issue; they bypass the specific reconstruction objectives of these input images and generate results that align well with the trained domain corresponding to the text.

\begin{figure}[t]
\centering
\vspace{-5px}
\includegraphics[width=\linewidth]{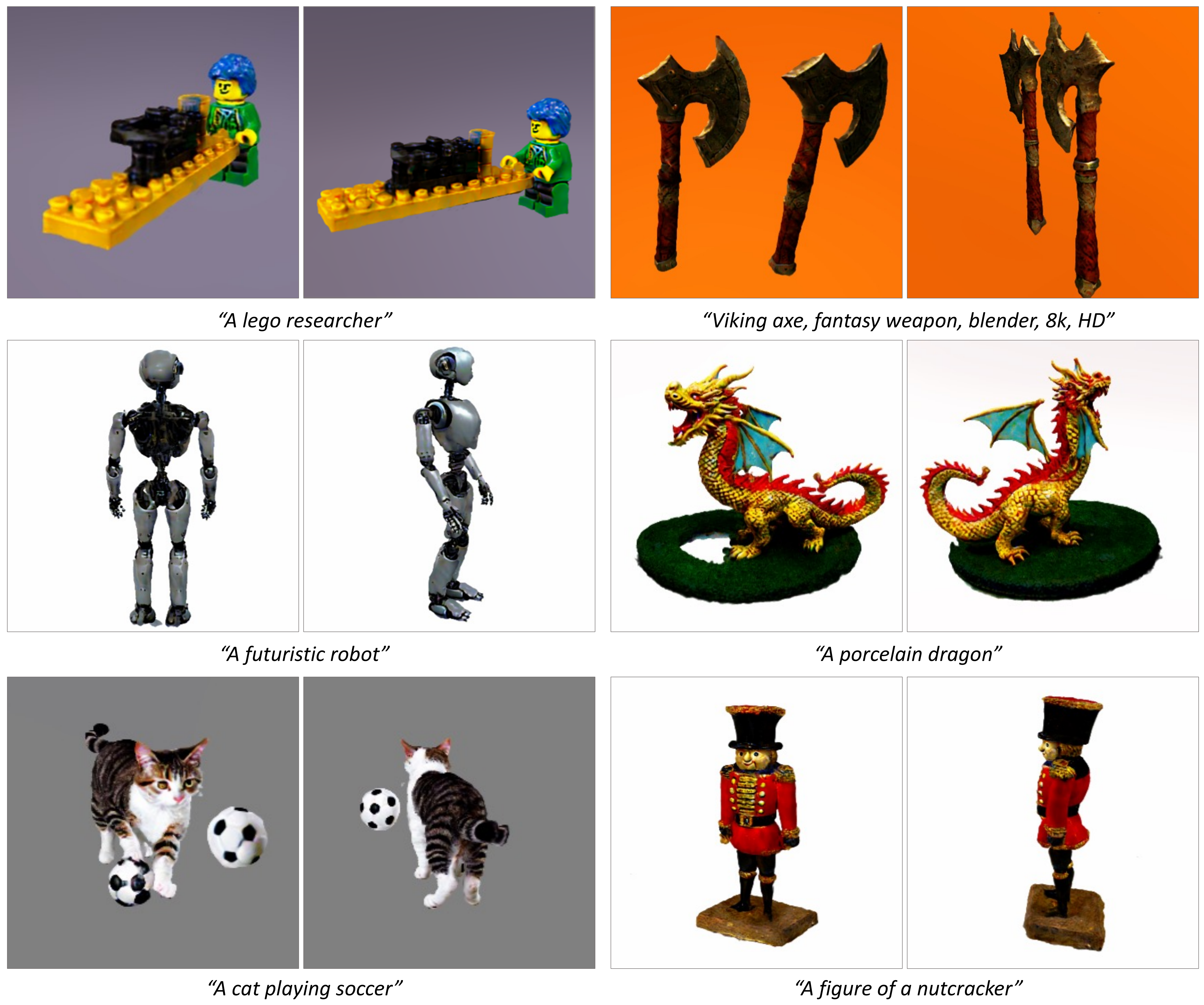}
\caption{\textbf{Additional qualitative results.} }
\label{fig:qual}
\vspace{-5px}
\end{figure}

\section{Additional Discussion}
\label{sec:discussions}

\subsection{Additional qualitative results.}
\label{sec:addqual_app}
We present additional qualitative results of our approach in Fig.~\ref{fig:qual}.

\subsection{Ablation on lightweight adaptation}
\label{sec:ablation_light_app}
We ablate the components of lightweight adaptation in Fig.~\ref{fig:abl_light}. In (a) of Fig.~\ref{fig:abl_light}, the 2D prior model is adapted using only the learnable layers that are embedded within the U-Net. In (b), both the learnable layers and tokens that correspond to the view prefix are adapted. To clearly demonstrate the differences, we present samples by deterministic DDIM~\cite{song2020denoising} sampler. We maintain consistency by using the same initial noises for all the samples. We observe that both (a) and (b) effectively mitigate the viewpoint bias inherent in the 2D prior model, indicating that both are capable of guide the model to generate images that are less biased in terms of viewpoint. We also observe that the samples from (b) represent a more diverse range of viewpoints.

\begin{figure}[h]
\centering
\includegraphics[width=1.\linewidth]{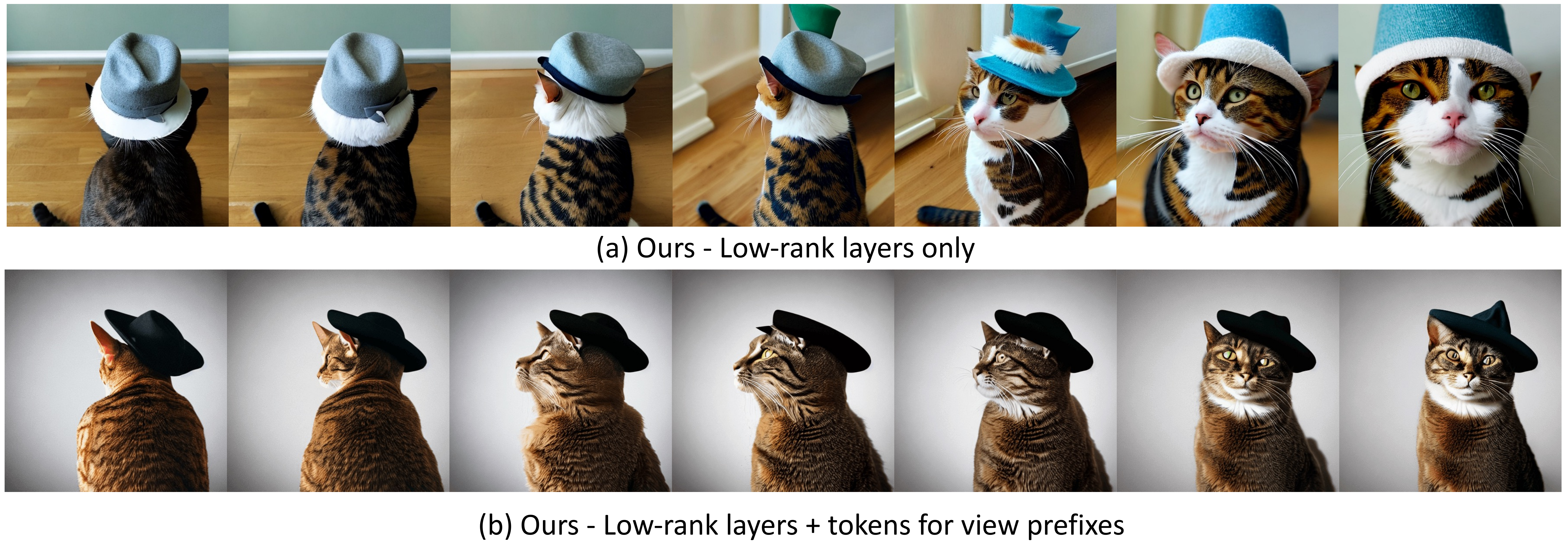}
\caption{\textbf{Ablation on components of lightweight adaptation.} (a): Adaptation using only the learnable layers embedded in the Diffusion U-Net. (b): Adaptation using the learnable layers and tokens corresponding to the view prefix. To clearly demonstrate the difference, we show samples from deterministic DDIM sampler after fixing the initial noise. In both (a) and (b), we observe that the viewpoint bias is effectively removed. However, in the case of (b), it shows samples from a slightly more diverse range of viewpoints.}
\label{fig:abl_light}
\end{figure}

\subsection{Does lightweight adaptation overfit the model to the retrieved asset?}
\label{sec:light_app}
In this section, we address the concern of potential overfitting to initializing assets during lightweight adaptation. To investigate this, we analyze 2D samples generated using a constant asset with progressively changing prompts. This lets us verify the level of overfitting our model display as it tunes itself to the specific details of the assets. Interestingly, as shown in Fig.~\ref{fig:light_overfit_test}, our findings indicate that the adaptation focuses more on general aspects, such as viewpoint, rather than specific details like texture. This suggests that lightweight adaptation avoids overfitting to the minor details of of individual assets, striking a balance between adapting to the 3D asset and maintaining generalization across various prompts.

\begin{figure}[t]
\center
\includegraphics[width=0.9\textwidth]{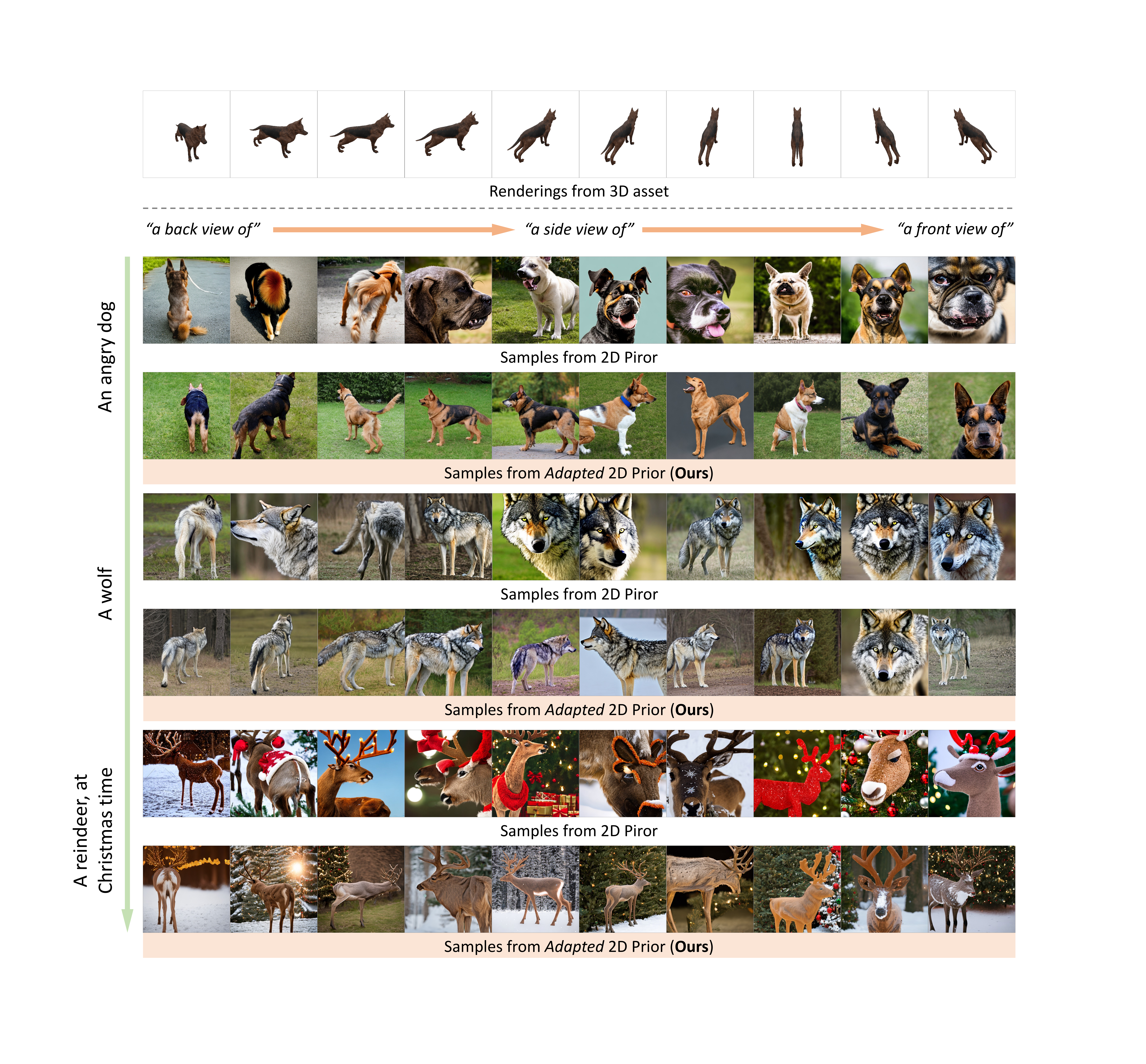}
\caption{To verify whether the model overfits to the retrieved assets for lightweight adaptation, we report on 2D samples generated by progressively changing prompts, while keeping asset same. It shows that the adaptation focuses more on general aspects like viewpoint, rather than specific details like texture. For more details, see Sec.~\ref{sec:light_app}.  }
\label{fig:light_overfit_test}
\end{figure}

\subsection{Ablation on each component}
\label{sec:ablation_app}
We conduct an ablation study on the two primary methodologies proposed in our main paper: initializing the variational distribution, and employing lightweight adaptation. As shown in Fig.~\ref{fig:abl}, initialization of the variational distribution is vital in solidifying coarse geometry, while lightweight adaptation shows its efficacy in preventing Janus problem-like artifacts.

\subsection{Orientation alignment}
\label{sec:or_align}
To verify whether retrieved 3D assets are oriented properly and well aligned to our canonical space axis, we measure the success rate of the alignment of assets retrieved for 45 prompts manually. To minimize human error in the measurement process, we follow these principles: 1) If the frontal view is correctly identified, it's a success, and it is considered a failure if its orientation is flipped vertically (upside-down), or flipped sideways, or if the frontal view cannot be identified. 2) We set a reference angle for the frontal view of each asset based on the horizontal axis, and define the failure case as occasions where the frontal view deviates by more than $\pm45$ degrees from the reference angle. 3) Radially symmetric objects or objects with semantic symmetry that whose frontal view cannot be identified singularly are excluded from the measurement. The results are reported in Tab.~\ref{table:orientation}. Note that failure cases here do not necessarily mean failures in generation, thanks to view prefix optimization in the adaptation.

\begin{table}[h]
\small
\centering
\vspace{5pt}
\begin{tabular}{ccc} 
\toprule
Success & Vertical / Sideways Inversion & Frontal view not identified \\
\midrule
86.7\% & 6.7\% & 6.7\%      \\
\bottomrule
\end{tabular}
\caption{\textbf{Success rate of orientation alignment.}}
\label{table:orientation}
\end{table} 

\subsection{Efficiency}
\label{sec:efficiency}
Our method is a retrieval-augmented approach that requires test-time adaptation. Unlike methods such as Zero123~\cite{} which involve tuning all of the parameters for 3D awareness with 1,344 GPU hours, our method does not require full fine-tuning in model preparation phase. The aspect of our method not requiring training is similar to DreamFusion~\cite{} or ProlificDreamer~\cite{}.

Instead, in inference time, our method includes a few more steps than classic SDS-based methods~\cite{}; 3D retrieval and lightweight adaptation. The 3D retrieval process takes about 7 seconds. The lightweight adaptation, when measured separately, takes about 7 minutes. However, as it actually proceeds in parallel with the SDS optimization process, the time taken is expected to be less from a total time perspective. 

SDS-based methods generate 3D objects through optimization, making it difficult to report exact generation times. This is different from optimization-based methods for 3D reconstruction~\cite{chen2022tensorf, muller2022instant, yu2021plenoxels} that report time based on reaching a certain PSNR, as it's challenging to know when it is converged. Consequently, we present qualitative results in Fig.~\ref{fig:converge}, the intermediate rendering results at 10,000th iteration. The results show that our method reaches convergence faster than our baseline~\cite{wang2023prolificdreamer}.

Based on this observation, we converge 3D objects over 20,000 iterations, which is 5,000 fewer iterations than the baseline. Therefore, the average time for generation ultimately becomes about 2 hours faster than the baseline.

\begin{figure}[t]
\centering
\vspace{-5px}
\includegraphics[width=0.9\linewidth]{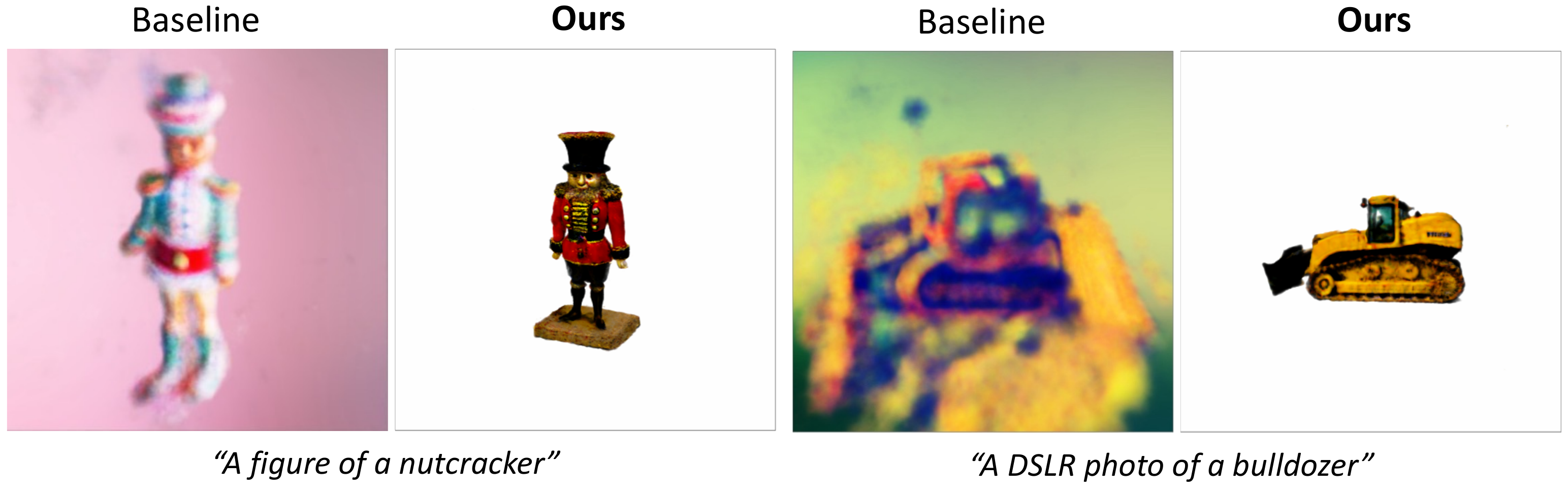}
\caption{\textbf{Intermediate renderings at 10,000th iteration.} }
\label{fig:converge}
\vspace{-5px}
\end{figure}

\subsection{Number of particles}
\label{sec:particles_app}
In this section, we present an ablation study on the number of particles, as shown in Fig.~\ref{fig:particles}. Similar to the findings in the ablation study for VSD, as described in Appendix E.3 of \cite{wang2023prolificdreamer}, it has been observed that the impact on quality due to the number of particles is not significant. We have found that there is a tendency for increased diversity with a greater number of particles. This appears to be due to the fact that as the number of particles increases, the ability to absorb a larger number of retrieved assets for distribution initialization becomes more effective, thus enhancing diversity. In the perspective of 3D consistency, even with a smaller number of particles, the number of retrieved assets that can be used in lightweight adaptation is not limited, hence a slight improvement in performance or a similar trend has been observed.

\section{Applications}
\label{sec:applications}
 Our retrieval-based text-to-3D generation method can be applied to numerous real-life cases in which 3D models is necessary. In most cases, creating a realistic 3D model requires extensive knowledge of complicated tools and programs, such as CAD, which limits a layman from creatively engaging in 3D scene generation. Our retrieval-based text-to-3D generation method enables the score distillation-based optimization process to be controlled by both text and retrieved asset, giving higher flexibility and diversity to 3D scenes that can be generated. This opens up large possibilities in all areas which requires 3D creation: in AR and gaming. Our retrieval-based methodology can be used to design 3D models of characters or buildings that are more meticulously generated under user control, greatly reducing the redundant time and cost that goes into crafting such 3D assets with hand. Due the realism and fidelity of generated 3D scenes, our model can also be applied to aid 3D design that goes into movies for CGI-based scenes, giving artists more relevant 3D mesh that serves as more efficient template from which they can work and fine-tune upon.

\paragraph{3D data enhancement.} Furthermore, our method can also be applied to to specifically enhance the fidelity and details of low-quality 3D assets by simply replacing retrieved asset with the 3D asset to be enhanced. As demonstrated in Sec.~\ref{sec:discussions}, we show that high-quality 3D model that preserve general geometric structure of the given asset can be created when the assets are not automatically retrieved by hand-picked as input to the given network, despite the low quality of initializing assets. Note also that even when the texts and assets themselves not being completely aligned semantically, (e.g., 3D asset of a plain human figure and text prompt ``\textit{A photo of Ironman}''), our model successfully enhances the asset in accordance with the text prompt. This show that our work can be extended to 3D data enhancement in settings where the assets are hand-picked and chosen to be improved by text prompt. 
\begin{figure}[t]
\centering
\vspace{-5px}
\includegraphics[width=0.6\linewidth]{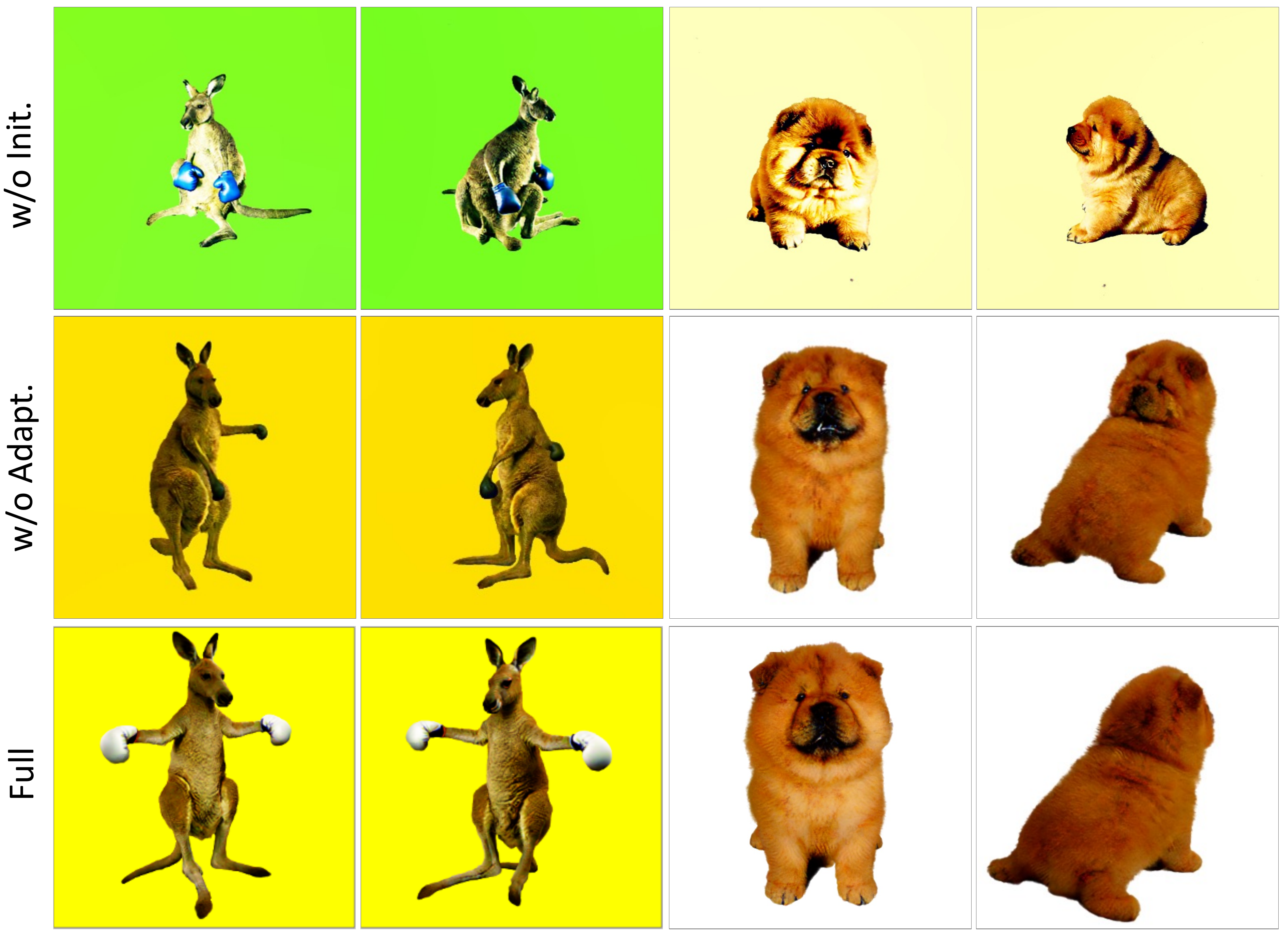}
\caption{\textbf{Ablation on each component.} We drop each component of our framework. Top row and middle row show the results generated by dropping out the initialization of the distribution and lightweight adaptation respectively. The results in the bottom row is produced by our whole framework and they show more consistent geometry compared to upper rows. }
\label{fig:abl}
\vspace{-5px}
\end{figure}

\section{Details of User Study}
\label{sec:user_study}
The user study involved a total of 92 participants, each of whom was asked to answer 6 randomly sampled questions. Specifically, each question presented two videos showing our results and baseline's results. It was thoroughly concealed which video is the baseline and which is our result, and the placement of the videos was also randomized. The questions are as follows:
\begin{itemize}
    \item The text used for this 3D creation is ``[TEXT PROMPT]''. Considering \textbf{texture, shape, geometry}, which result do you find more satisfactory?
\end{itemize}
This questionnaire was distributed for 3 days in local communities and universities, and stakeholders in this study were strictly excluded, and which result come from which model's results was also strictly blinded. We provide an example of the screen shown to the participants in Fig.~\ref{fig:userstudy}.

\section{Limitations}
\label{sec:limitations}
Our generation process approximately takes about 6 hours, which is faster than our implementation baseline, ProlificDreamer~\cite{wang2023prolificdreamer} which takes about 8 hours. However, it should be noted that, it takes longer time than concurrent works which concentrate on fast inference~\cite{yi2023gaussiandreamer}, as our goal is to create photorealistic 3D contents like ProlificDreamer, not to make the infernece faster. We believe it would be possible to significantly reduce the time required by applying these techniques in an orthogonal manner as a future work.

Secondly, the receptivity of complex and creative text prompts is bounded by the performance of the 2D prior model, Stable Diffusion~\cite{ldm}. In this paper, while we utilize Stable Diffusion 2.1 for a fair comparison with other work~\cite{wang2023prolificdreamer}, it should be noted that the recent advancements in 2D generative models suggest methods for a better understanding of more complex prompts, which could be pursued in our future work.


\begin{figure}[t]
\centering
\includegraphics[width=0.9\linewidth]{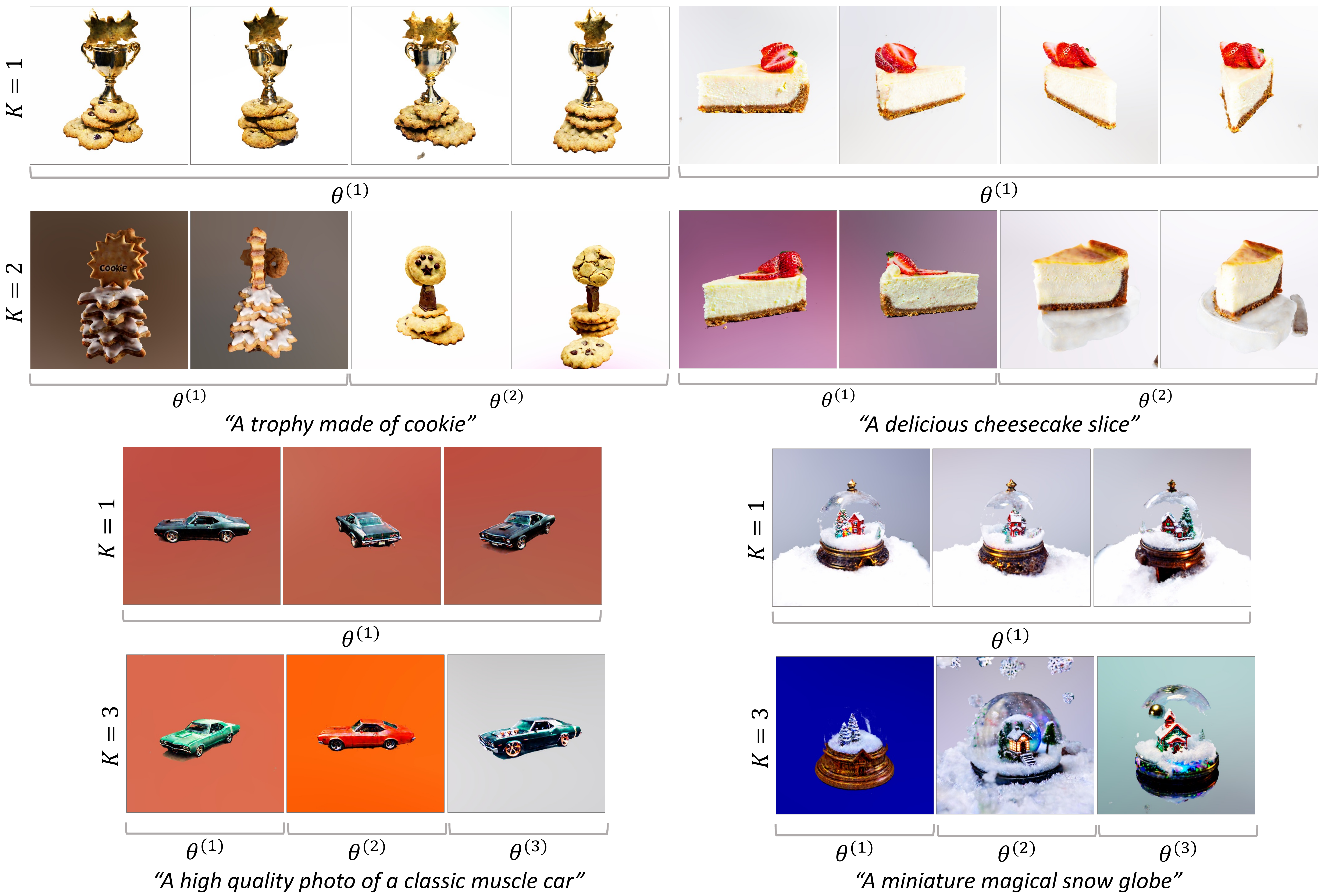}
\caption{\textbf{Variation of the number of particles.} We show the generation results with different number of particles. }
\label{fig:particles}
\vspace{-5px}
\end{figure}


\begin{figure}[h]
\centering
\vspace{-5px}
\includegraphics[width=0.4\linewidth]{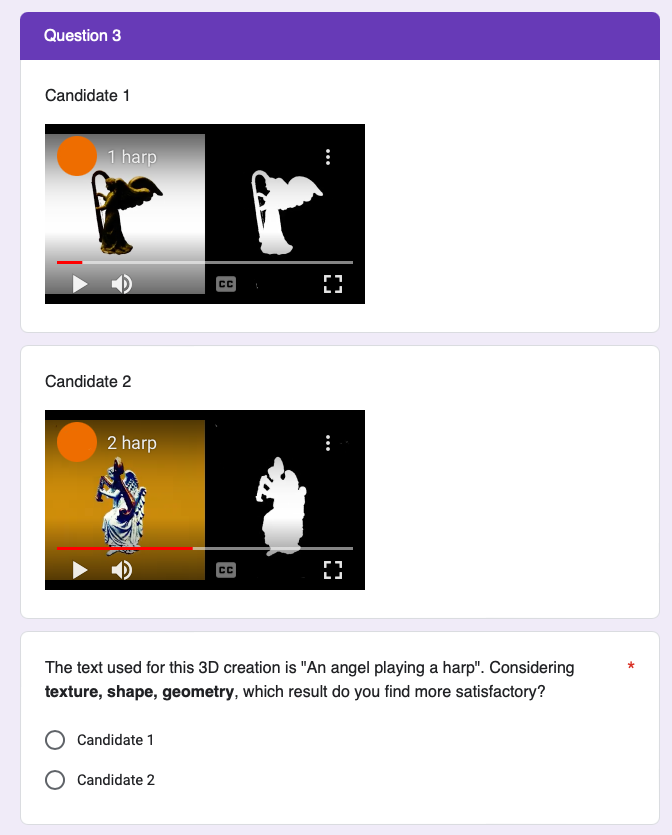}
\caption{\textbf{An example of the screen shown to participants.} Some contents are obscured in this example. }
\label{fig:userstudy}
\vspace{-5px}
\end{figure}


\end{document}